\documentclass[letterpaper, 10 pt, conference]{ieeeconf} 

\IEEEoverridecommandlockouts 

\usepackage{times} 
\usepackage{amsmath} 
\usepackage{amssymb} 
\usepackage{graphicx}
\usepackage{cases}
\usepackage{algorithm}
\usepackage{algorithmic}
\usepackage{caption}
\usepackage{color}
\usepackage{subfigure}
\usepackage{cite}

\title{\LARGE \bf
Deep reinforcement learning of event-triggered communication and consensus-based control for distributed cooperative transport
}

\author{Kazuki Shibata$^{1}$, Tomohiko Jimbo$^{1}$ and Takamitsu Matsubara$^{2}$
\thanks{$^{1}$Toyota Central R$\&$D Labs., Inc., 41-1, Yokomichi, Nagakute, Aichi, 480-1192, Japan.
{\tt\small kshibata@mosk.tytlabs.co.jp}}%
\thanks{$^{2}$Division of Information Science, Graduate School of Science and Technology, Nara Institute of Science and Technology, 8916-5, Takayama-cho, Ikoma, Nara, 630-0101, Japan
{}}%
\thanks{$^{}$\it{}Preprint submitted to Elsevier on 14th November, 2022
{}}%
}

\begin{document}

\maketitle
\thispagestyle{empty}
\pagestyle{empty}

\begin{abstract}
In this paper, we present a solution to a design problem of control strategies for multi-agent cooperative transport. Although existing learning-based methods assume that the number of agents is the same as that in the training environment, the number might differ in reality considering that the robots' batteries may completely discharge, or additional robots may be introduced to reduce the time required to complete a task. Therefore, it is crucial that the learned strategy be applicable to scenarios wherein the number of agents differs from that in the training environment. In this paper, we propose a novel multi-agent reinforcement learning framework of event-triggered communication and consensus-based control for distributed cooperative transport. The proposed policy model estimates the resultant force and torque in a consensus manner using the estimates of the resultant force and torque with the neighborhood agents. Moreover, it computes the control and communication inputs to determine when to communicate with the neighboring agents under local observations and estimates of the resultant force and torque. Therefore, the proposed framework can balance the control performance and communication savings in scenarios wherein the number of agents differs from that in the training environment. We confirm the effectiveness of our approach by using a maximum of eight and six robots in the simulations and experiments, respectively.
\end{abstract}

\section{Introduction}
Cooperative transport is an important research topic in robotics and can be applied in fields such as warehouse logistics \cite{DAndrea2012, Barrientos2011} and search and rescue missions \cite{Queralta2020}. Multi-agent systems are particularly advantageous for transporting a payload that cannot be moved by a single agent.
Moreover, these systems can manage scenarios wherein robots experience actuator failures or battery-power depletion.

In this study, we address a design problem of control strategies for multi-agent cooperative transport.
Most previous studies on multi-agent cooperative transport have used wireless communication to share observations among agents. However, if multiple robots transmit information at high fixed rates in the same network system, the communication bandwidth can be compressed. This will increase the probability of message loss and cause long transmission delays \cite{Zhang2016}. Therefore, it is crucial to minimize communication.  Previous studies~\cite{Franchi2014,Franchi2015,Petitti2016,Culbertson2018} have employed distributed adaptive/robust control to deal with the unknown object dynamics in the multi-robot cooperative transport. However, these studies have strong
assumptions and can only work with simple tasks where agents are rigidly attached to the payload. Therefore, it is crucial to consider a data-driven framework that does not require strong assumptions.

In this study, we explored a multi-agent reinforcement learning (MARL) approach \cite{Lowe2017, Foerster2018} to simultaneously solve the design problems of communication and control strategies for multi-agent cooperative transport.
To date, several studies have proposed communication strategies to reduce the communication frequency \cite{Baumann2018, Funk2020}, as well as the number of communicating agents and data \cite{Shibata2021}.
These studies assume that the number of agents is equivalent to that in the training environment. However, in reality, the number of robots can differ owing to the discharging of the robots' batteries, or additional robots may be introduced to complete tasks quickly. Therefore, the learned strategy must also be applicable to scenarios wherein the number of agents differs from that in the training environment.

\begin{figure}
\hspace{1mm}
\includegraphics[width=8cm]{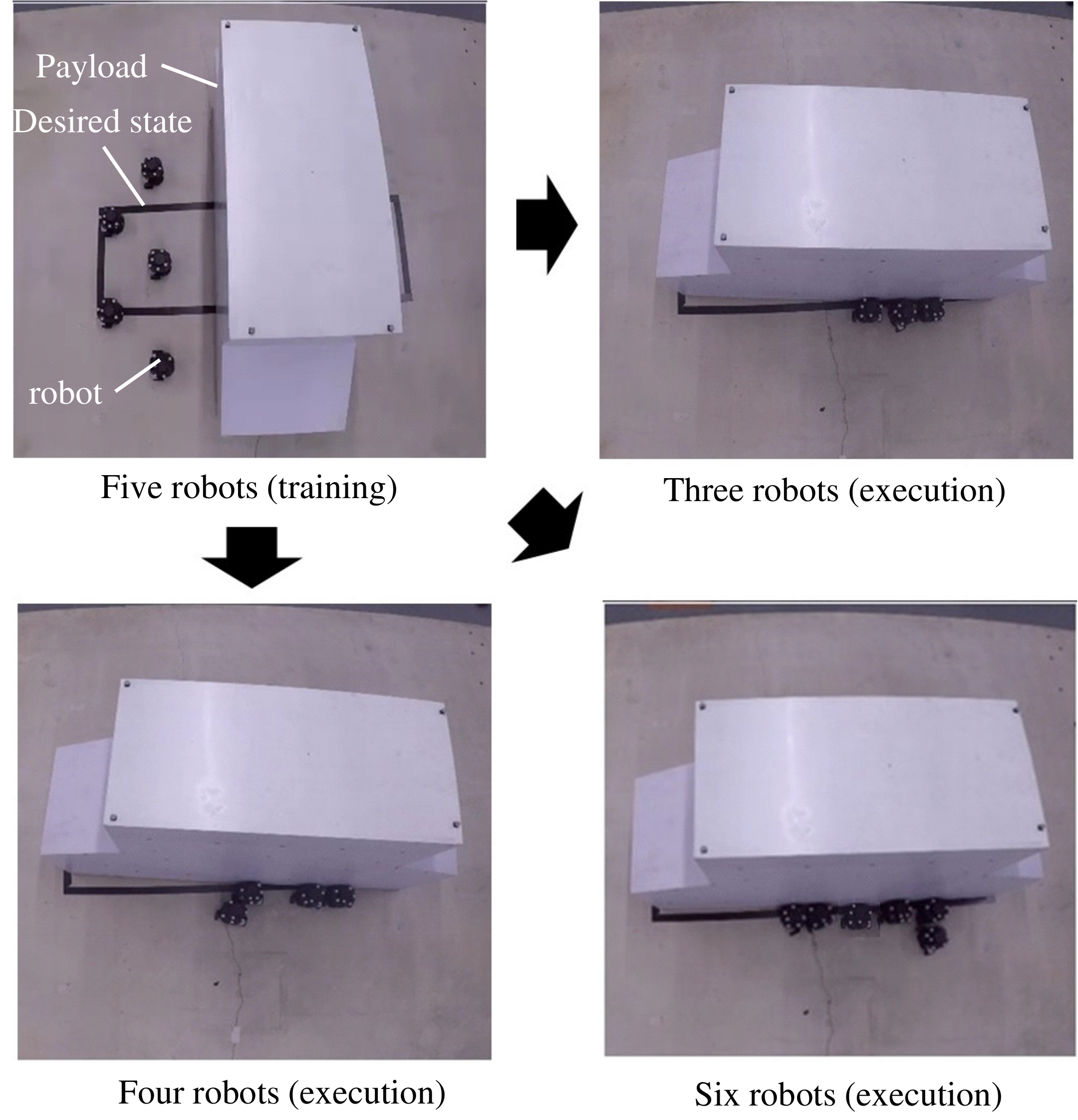}
\caption{Cooperative rotation task using multiple robots. The proposed framework can control the payload to the desired state for scenarios wherein the number of robots differs from that in the training environment.}
\label{intro}
\end{figure}

The objective of the study was to control the payload to the desired state for scenarios wherein the number of robots differs from that in the training environment, as shown in Fig. \ref{intro} while minimizing communication among robots.
To achieve this, we propose a novel MARL framework of event-triggered communication and consensus-based control for distributed cooperative transport. For the settings in this research, the numbers of communication and environmental agents were maintained constant during training but variable during execution. The proposed policy model enables each agent to establish agreements on global information with local communication agents. Limiting the minimum number of environmental agents to a fixed number of communication agents enables the proposed method to be applicable to a varying number of agents. In particular, our policy model estimates the resultant force and torque applied to an object in a consensus manner \cite{Saber2004, Ren2005} using the estimates of the resultant force and torque of two communication robots.
Moreover, under local observations and estimates of the resultant force and torque, the proposed framework can compute the control and communication inputs to determine when to communicate with the neighboring agents.
Therefore, the control performance and communication savings can be balanced, despite the number of agents being different from that in the training environment.

Although our preliminary study \cite{Shibata2021} used a communication strategy similar to that in the present study, it could not estimate the global information using the local information. Therefore, our preliminary study failed in tasks that could not be solved using only local information; however, our current study succeeds in this regard, as demonstrated by numerical simulations.
To the authors' knowledge, no similar studies on learning event-triggered control of multi-agent systems have been reported so far.

We demonstrate the effectiveness of the proposed algorithm through cooperative transport and cooperative rotation tasks.
We confirm the versatility of our framework through cooperative transport task using two agents for randomly arranged initial and desired positions of the payload in the simulations.
Moreover, we confirm the scalability of our framework by using a maximum of eight and six robots in the simulations and experiments, respectively.

The contributions of this study are as follows:
\begin{itemize}
\item We propose a learning framework of event-triggered communication and consensus-based control for distributed cooperative transport.
\item Unlike the distributed adaptive/robust control, our method does not require the dynamics of cooperative transport and can be applied to a wide range of tasks and not exclusively to tasks in which agents are rigidly attached to a payload.
\item We confirm that the proposed framework can balance control performance and communication savings in scenarios wherein the number of agents differs from that in the training environment.
\item We confirm that our algorithm can control the payload for varying number of agents through real robot experiments.
\end{itemize}

The remainder of this paper is organized as follows.
Section II introduces related works. Section III describes preliminary on the distributed cooperative transport and consensus problems.
Section IV introduces the MARL setting for distributed cooperative transport and the proposed framework of event-triggered communication and consensus-based control.
Section V demonstrates the effectiveness of our algorithm through numerical simulations.
Section VI demonstrates the effectiveness of our method through real robot experiments.
Section VII discusses the limitations and scope of future work.
Finally, Section VIII presents the conclusions of this study.

\section{Related work}
In this section, we introduce model-based approaches that derive the control policy using distributed control, based on the dynamics model of cooperative transport. Furthermore, model-free approaches are introduced, which derive the control policy using data-driven approaches without the dynamics model.

\subsection{Model-based approaches} 
Previous studies on cooperative transport have derived control strategies based on the dynamics model. Furthermore, they have successfully
demonstrated various tasks using multiple arm robots \cite{WangICRA2016, WangIJRR2016}, human and robots \cite{Sieber2015, Gienger2018}, and aerial manipulation by multiple quadcopters using cables \cite{Michael2011, Jiang2013, Sreenath2013} or electromagnetic grippers \cite{Mellinger2013, Loianno2018}.

A key issue of cooperative transport lies in controlling the payload without prior knowledge of the payload and robots.
Franchi et al. \cite{Franchi2014, Franchi2015} proposed a decentralized parameter estimation of an unknown load using observations of the neighboring agents.
Based on this algorithm, Petitti et al. \cite{Petitti2016} proposed a robust control strategy to stabilize the payload in the presence of estimation uncertainties. 
Marino et al. \cite{Marino2018} proposed a distributed control strategy with the estimation of an unknown object without explicit communication and prior knowledge of the number of robots.
Culbertson et al. \cite{Culbertson2018} proposed a distributed adaptive control strategy for cooperative transport of an unknown object without parameter estimation.
Their control strategy required no communication between agents and made the payload state asymptotically converge to the desired state with a theoretical proof using the Lyapunov function.

Other studies employed variable-rate communication to reduce the communication frequency.
Dimarogonas et al. \cite{Dimarogonas2012} introduced event-triggered control \cite{Heemels2012, Miskowicz2016} into multi-agent communication to determine the timing of the communication with neighboring agents.
Trimpe et al. \cite{Trimpe2011} proposed a distributed control strategy with event-triggered communication to determine both the timing and transmitted data based on the error between the actual measurements and the estimates.
Furthermore, they demonstrated the effectiveness by conducting balancing cube experiments using six modules, each having sensors, actuation, and computational units.
Dohmann et al. \cite{Dohmann2020} were the first to propose a distributed control strategy with event-triggered communication for cooperative manipulation. Their methods minimized the frequency of receiving positions and velocities of end effectors from neighboring agents while accomplishing several manipulation tasks.

However, these approaches require a dynamics model and cannot be applied to tasks wherein dynamics models are difficult to formulate. In contrast, the proposed method is model-free and can be applied to more cooperative transport tasks compared to these approaches.

\subsection{Model-free approaches}
Several recent studies \cite{Rahimi2018, Miyazaki2021, Niwa2021} have adopted MARL approaches for multi-agent cooperative transport without requiring a dynamics model.
However, one of the main problems in MARL is that the variance of the estimated policy becomes large owing to the changing policies of other agents \cite{Hernandez2019}.
To address this issue, several authors \cite{Lowe2017, Foerster2018} proposed a learning framework of centralized training and decentralized execution, which learns critics for multi-agents and derives a decentralized policy using observations from each agent;
however, these methods cannot determine the timing of communication and can only function with fixed-rate communication.

To learn control strategies while saving communication costs, several authors \cite{Baumann2018, Funk2020} proposed a policy model using event-triggered control to minimize the control signals from a single learning agent to its actuator while achieving the control objective.
Demirel et al. \cite{Demirel2018} proposed DEEPCAS, a reinforcement learning-based control-aware scheduling algorithm in multi-agent setups. In this method, a centralized scheduler called DEEPCAS allocates $M$ communication channels to $N$ agents ($M\ll N$) while designing the agents' controllers beforehand.
Our preliminary study \cite{Shibata2021} extended the policy model \cite{Baumann2018, Funk2020} in multi-agent setups to reduce the frequency, number of communicating agents, and transmitted data. Although these methods could save communication costs, they adopted a policy model with inputs dependent on the number of agents and were inapplicable to problems wherein the number of agents was different from that in the learning environment.

The proposed framework combines the estimation of the resultant force and torque in a consensus manner and an event-triggered communication to determine the timing of communication into a policy model. Considering that our policy model computes the control and communication inputs under local observations and the estimates of the resultant force and torque of the neighborhood agents, it can be applied to scenarios wherein the number of agents differs from that in the training environment.

While the present study adopts a policy model using event-triggered control, it differs from the methodology proposed in \cite{Baumann2018, Funk2020, Shibata2021} as it involves an estimation mechanism and a distributed policy model under local observation. Moreover, the proposed framework can transport the payload to the desired state for varying number of agents.

\section{Preliminary}
\subsection{Distributed cooperative transport problem}

\subsubsection{Notation}
Let the position, yaw angle, velocity, angular velocity, desired position, and the desired yaw angle of the payload in world coordinates be denoted by $\textit{\textbf{x}}\in \mathbb{R}^2$, $\theta \in \mathbb{R}$, $\textit{\textbf{v}}\in \mathbb{R}^2$, $\omega\in \mathbb{R}$, $\textit{\textbf{x}}^{\ast} \in \mathbb{R}^2$, and $\theta^{\ast} \in \mathbb{R}$, respectively.

The position, yaw angle, and control input of the agent $i$ ($i=1,\cdots,N$) are represented by $\textit{\textbf{x}}_i\in \mathbb{R}^2$, $\theta_i \in \mathbb{R}$, and $\textit{\textbf{u}}_i\in \mathbb{R}^2$, respectively.
Agent $i$ applies a force $\textit{\textbf{f}}_i \in\mathbb{R}^2$ and torque $\tau_i \in\mathbb{R}$ on the payload.

\subsubsection{Problem formulation}
We consider a team of $N$ agents pushing a rigid payload with an unknown mass and moment of inertia. In our setting, $N$ is constant during the training phase but variable during the execution phase.
The objective of this problem is to control the payload to its desired state while reducing communication with other agents for varying numbers of agents during the execution phase.

According to \cite{Culbertson2018}, we assumed the following:
\begin{itemize}
\item all agents know $\textit{\textbf{x}}^{\ast}$ and $\theta^{\ast}$;
\item agent $i$ can observe $\textit{\textbf{x}}$, $\theta$, $\textit{\textbf{v}}$, $\omega$, $\textit{\textbf{x}}_i$, $\theta_i$, $\textit{\textbf{f}}_i$, and $\tau_i$.
\end{itemize}

\begin{figure}
\begin{center}
\includegraphics[width=6.3cm]{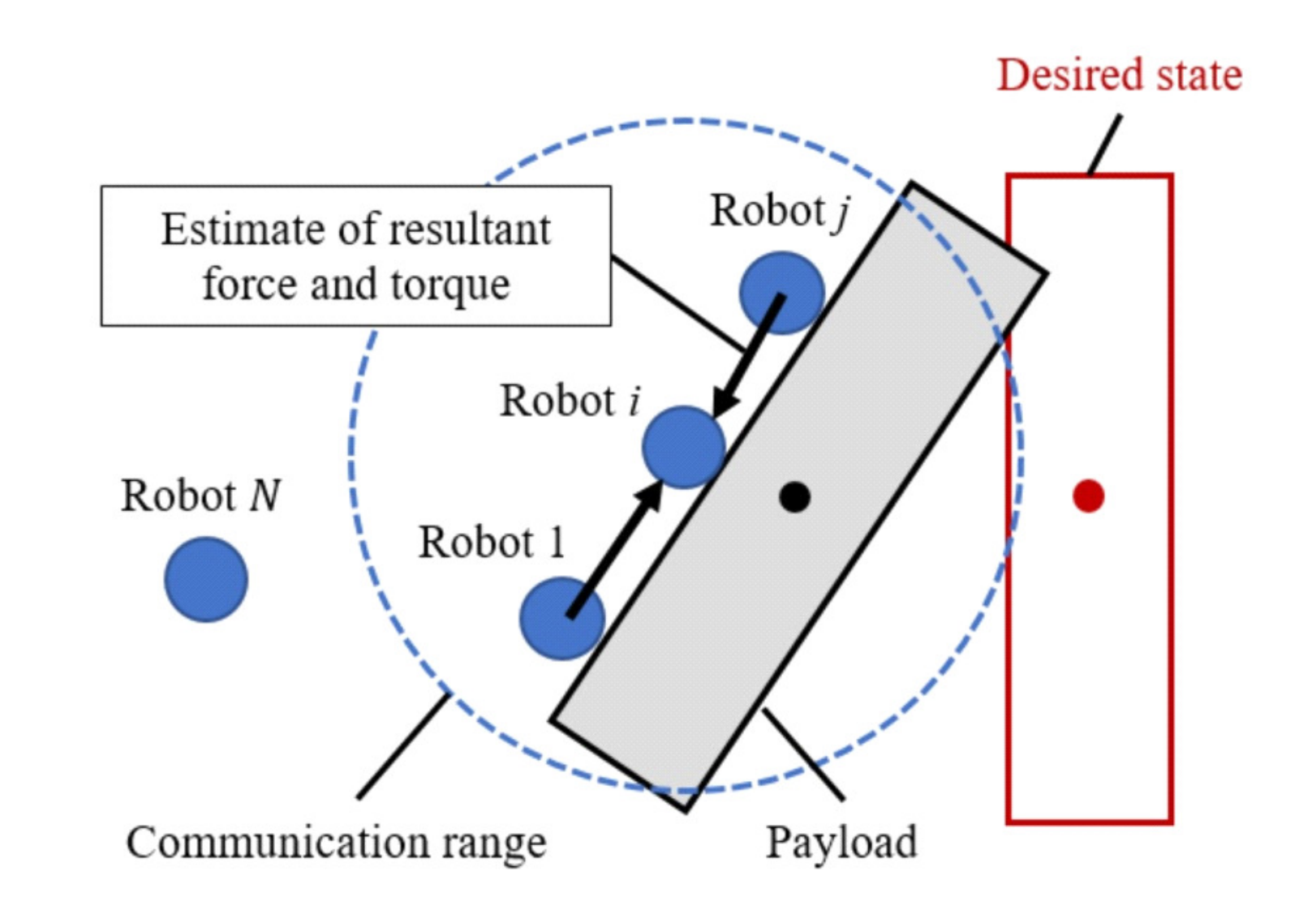}
\caption{Distributed cooperative transport.}
\label{setting}
\end{center}
\end{figure}

Moreover, we assumed the following:
\begin{itemize}
\item all agents know $N$.
\item the agent can communicate the estimates of the resultant force and torque with the $K$ nearest agents, as shown in Fig. \ref{setting}, where $K$ is constant.
\end{itemize}

\subsection{Consensus problem}
\subsubsection{Notation}
Let a binary variable be defined by $\gamma_{ij}$; using this, agent $i$ receives data from agent $j$. Specifically, $\gamma_{ij}=1$ if agent $i$ receives data from agent $j$; otherwise, $\gamma_{ij}=0$.

The entries of the adjacency matrix $\textit{\textbf{A}}\in \mathbb{R}^{N\times N}$, i.e., $A_{ij}$ $(i,j\in\{1,\cdots,N\})$, are given by
\begin{eqnarray}
A_{ij}\leftarrow
\begin{cases}
1,\ {\rm if}\ \gamma_{ij}=1 \\
0,\ {\rm otherwise}
\end{cases}.
\nonumber
\end{eqnarray}

The degree matrix $\textit{\textbf{D}}\in \mathbb{R}^{N\times N}$ is a diagonal matrix whose entries $D_{ij}$ $(i,j\in\{1,\cdots,N\})$ are given by
\begin{eqnarray}
D_{ij}\leftarrow
\begin{cases}
d_i,\ {\rm if}\ i=j \\
0,\ {\rm otherwise}
\end{cases},
\nonumber
\end{eqnarray}
where $d_i$ represents the total number of agents that communicate with agent $i$.

\subsubsection{Communication topology}
We define several terms related to the communication topology based on graph theory. The communication topology is undirected if the communication between agents is bi-directional; otherwise, the communication topology is directed. Moreover, the communication topology is connected if communication is possible for any agent when starting from any agent at adjacent agents; otherwise, the communication is disconnected.

According to \cite{Saber2007}, the communication topology is connected if the following condition is satisfied.
\begin{eqnarray}
{\rm rank}(\textit{\textbf{L}})=N-1,
\label{rank}
\end{eqnarray}
where $\textit{\textbf{L}}:=\textit{\textbf{D}}-\textit{\textbf{A}}$ is
the graph Laplacian. 

\subsubsection{Problem formulation}
We consider $N$ agents with a vector $\textit{\textbf{c}}:=[\textit{\textbf{c}}_1,\cdots,\textit{\textbf{c}}_N]$. The objective of this problem is to converge $N$ vectors to the same value. One common method is Laplacian averaging \cite{Elhage2010}, which is used to average the estimates of $N$ agents. This algorithm achieves a consensus given by
\begin{eqnarray}
\textit{\textbf{c}}[s+1]=\textit{\textbf{c}}[s]-k \textit{\textbf{L}}\textit{\textbf{c}}[s],
\label{Laplacianaveraging}
\end{eqnarray}
where $\textit{\textbf{c}}[s]$ represents the vector at step $s$, and $k$ is a positive constant.
Using Eq. (\ref{Laplacianaveraging}), we can make $\textit{\textbf{c}}$ converge to the average value after $m$ iterations, as follows:
\begin{eqnarray}
\lim_{m\rightarrow \infty}\textit{\textbf{c}}[s+m]=\frac{\textbf{1}\textbf{1}^{\top}}{N}\textit{\textbf{c}}[s]
\label{averageconsensus}
\end{eqnarray}

According to \cite{Kennedy2015}, the consensus in Eq. (\ref{averageconsensus}) can be guaranteed if the following conditions are satisfied:
\begin{itemize}
\item The communication topology is undirected and connected.
\item $0<k<\frac{2}{N}$
\end{itemize}

\section{Deep reinforcement learning of event-triggered communication and consensus-based control}
This section introduces a MARL framework that can be applied to cooperative transport with varying numbers of agents.
We introduce the setting of MARL for distributed cooperative transport and the proposed learning method.

\subsection{Setting of MARL for distributed cooperative transport}
In what follows, we introduce the MARL setting for distributed cooperative transport according to a Markov decision process.

We denote the state, observation, and action of agent $i$ as $\textit{\textbf{s}}_i$, $\textit{\textbf{o}}_i$ and $\textit{\textbf{a}}_i$, respectively.
Agent $i$ selects action $\textit{\textbf{a}}_i$ under its local observation $\textit{\textbf{o}}_i$ depending on a policy $\pi_i$.
After $N$ agents select the current actions $[\textit{\textbf{a}}_1,\cdots,\textit{\textbf{a}}_N]$, the current states $[\textit{\textbf{s}}_1,\cdots,\textit{\textbf{s}}_N]$ transition to the next states $[\textit{\textbf{s}}'_1,\cdots,\textit{\textbf{s}}'_N]$.
At current step $t$, agent $i$ receives a reward $r_t\in \mathbb{R}$, which is defined by the error between the current and desired state of the payload and the communication costs.
Agent $i$ updates its policy by maximizing the expected reward $\mathbb{E}[R_t]=\mathbb{E}\left[\sum^{T-1}_{k=0}\gamma^{k}r_{t+k}\right]$, where $\gamma \in [0, 1]$ is a discount factor, and $T$ is the total number of control steps per episode.

\subsection{Distributed policy model}
\subsubsection{Overview}
\begin{figure}
\begin{center}
\includegraphics[width=8cm]{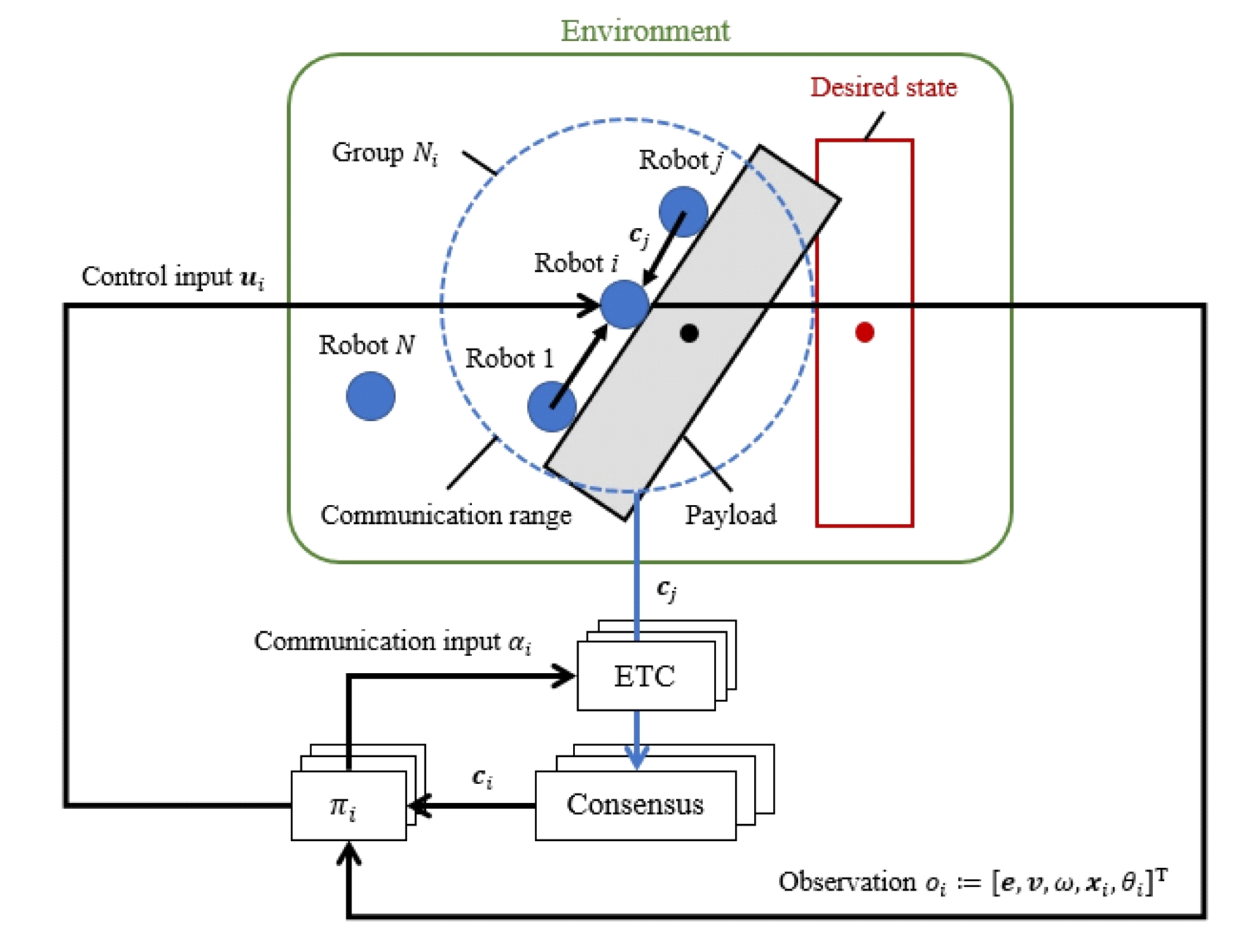}
\caption{The proposed policy model of event-triggered communication and consensus-based control for distributed cooperative transport}
\label{framework}
\end{center}
\end{figure}

Figure \ref{framework} presents the proposed policy model of event-triggered communication and consensus-based control for distributed
cooperative transport.
Our method exploits a distributed policy model that computes the communication and control inputs using local observations and the
resultant force and torque with consensus estimation.

Agent $i$ clusters $N$ agents into $K$ nearest agents in the group $\mathcal{N}_i$.
To ensure that the policy is scalable to the number of agents, each agent estimates the resultant force and torque using those of the neighborhood
agents.
Based on the consensus algorithm, agent $i$ estimates the resultant force $\textit{\textbf{F}}_i\in \mathbb{R}^2$ and torque $T_i\in \mathbb{R}$ using $\textit{\textbf{F}}_j$ and $T_j$ obtained from agent $j\in \mathcal{N}_i$ via communication.
Event-triggered communication (ETC) determines when to communicate with agent $j\in \mathcal{N}_i$ at every control step.
Policy $\pi_i$ calculates the control input $\textit{\textbf{u}}_i\in \mathbb{R}^2$ and communication input $\alpha_i\in \mathbb{R}$ using local
observation $\textit{\textbf{o}}_i=[\textit{\textbf{e}}^{\top},\textit{\textbf{v}}^{\top},\omega,\textit{\textbf{x}}_i^{\top}\theta_i]^{\top}$, $\textit{\textbf{F}}_i$ and $T_i$, where $\textit{\textbf{e}}=[\textit{\textbf{x}}^{\top}-{\textit{\textbf{x}}^{\ast}}^{\top},\theta-\theta^{\ast}]^{\top}$ is the error vector.

Considering that our policy model computes the control and communication inputs under local observations and the estimates of the resultant force and torque of the neighborhood agents, it can be applied to scenarios wherein the number of agents differs from that in the training environment.

\subsubsection{Consensus estimation}
This subsection details the estimation of the resultant force and torque in a consensus manner.
Let denote the estimates of agent $i$ by $\textit{\textbf{c}}_i=\frac{1}{N}[\textit{\textbf{F}}_i^{\top}, T_i]^{\top}$.
By communicating with nearest agents, $\textit{\textbf{c}}_i$ can be estimated by
\begin{eqnarray}
\textit{\textbf{c}}_i[t+\Delta t]=\textit{\textbf{c}}_i[t]+k \sum_{j\in \mathcal{N}_i}(\textit{\textbf{c}}_j[t]-\textit{\textbf{c}}_i[t]),
\label{consensus}
\end{eqnarray}
where $\Delta t$ is the consensus period.
At every control period $\Delta T$ ($>\Delta t$), agent $i$ updates $\textit{\textbf{c}}_i$ using $\textit{\textbf{c}}_i\leftarrow \left[\textit{\textbf{f}}_i^{\top}, \tau_i\right]^{\top}$.

Considering $\Delta t$ in Eq. (\ref{consensus}) is smaller than $\Delta T$, the communication costs may increase owing to the high-rate communication required for the estimation.
To address this issue, we introduce an event-triggered architecture that determines the timing of communication with neighborhood agents while controlling the payload to the desired state.

\subsubsection{Event-triggered communication and consensus-based control}
In this subsection, we introduce the ETC and consensus-based control that balances the transport performance and communication savings for varying number of agents.

In ETC, agent $i$ receives $\textit{\textbf{c}}_j$ from agent $j\in \mathcal{N}_i$ by
\begin{eqnarray}
\textit{\textbf{c}}_j=
\begin{cases}
\textit{\textbf{c}}_j,\ {\rm if}\ \gamma_{ij}=1 \\
0,\ {\rm if}\ \gamma_{ij}=0
\end{cases}.
\label{ETC}
\end{eqnarray}

According to \cite{Baumann2018, Funk2020}, the timing of communication is decided based on a trigger law given by
\begin{eqnarray}
\gamma_{ij}=1 \Longleftrightarrow \alpha_i \ge0
\label{ETClaw}
\end{eqnarray}
where $\alpha_i\in [-1,1]$ is an output of the policy of agent $i$. Using Eq. (\ref{ETClaw}), agent $i$ makes the communication decision with agent $j\in \mathcal{N}_i$ by
\begin{eqnarray}
\gamma_{ij}=
\begin{cases}
1,\ {\rm if}\ \alpha_i \ge 0 \ {\rm and} \ j\in \mathcal{N}_i\\
0,\ {\rm otherwise}
\end{cases},
\label{comdec}
\end{eqnarray}

Moreover, to ensure that the communication topology is undirected, agent $i$ updates $\gamma_{ij}$, given as follows:
\begin{eqnarray}
\gamma_{ij}\leftarrow
\begin{cases}
1,\ {\rm if}\ \gamma_{ji}=1 \\
\gamma_{ij},\  {\rm otherwise}
\end{cases}.
\label{update}
\end{eqnarray}

Our ETC and consensus-based control involves a policy that calculates the communication and control inputs using local
observations, as well as the resultant force and torque with consensus estimation, given as follows:
\begin{eqnarray}
\textit{\textbf{a}}_i=\left[\textit{\textbf{u}}_i^{\top},\alpha_i\right]^{\top}=\pi_i\left(\textit{\textbf{o}}_i, \textit{\textbf{c}}_i\right),
\label{policy}
\end{eqnarray}
where $\pi_i$ is computed by a deep neural network.
The communication input $\alpha_i$ is used to make the communication decision with other agents in the next control step using Eq. (\ref{comdec}).

The calculation steps used in the ETC and consensus-based control is shown in Algorithm \ref{algorithm}.

\begin{algorithm}
\caption{: ETC and consensus-based control}
\label{algorithm}
\begin{algorithmic}[1]
\STATE Initialize $\textit{\textbf{x}}$, $\theta$, $\textit{\textbf{x}}^{\ast}$, $\theta^{\ast}$, $\textit{\textbf{x}}_i$, $\theta_i$ ($i=1,\cdots,N$)
\STATE \textbf{for} $t=1$ to $T$ \textbf{do}
\STATE \ \ \ \ \ \ \textbf{for} $i=1$ to $N$ \textbf{do}
\STATE \ \ \ \ \ \ \ \ \ \ \ \ /* procedure for grouping $\mathcal{N}_i$
\STATE \ \ \ \ \ \ \ \ \ \ \ \ Calculate $K$ nearest agents among $N$ agents
\STATE \ \ \ \ \ \ \ \ \ \ \ \ /* procedure for ETC
\STATE \ \ \ \ \ \ \ \ \ \ \ \ Determine the agents to communicate using \\
\ \ \ \ \ \ \ \ \ \ \ \ Eqs. (\ref{comdec}) and (\ref{update})
\STATE \ \ \ \ \ \ \ \ \ \ \ \ /* procedure for consensus-based control
\STATE \ \ \ \ \ \ \ \ \ \ \ \ Update $\textit{\textbf{c}}_i\leftarrow [\textit{\textbf{f}}_i^{\top}, \tau_i]^{\top}$
\STATE \ \ \ \ \ \ \ \ \ \ \ \ \textbf{for} $s=t$ to $t+\Delta T$ \textbf{do}
\STATE \ \ \ \ \ \ \ \ \ \ \ \ \ \ \ \ \ \ Estimate $\textit{\textbf{c}}_i$ using Eqs. (\ref{consensus}) and (\ref{ETC})
\STATE \ \ \ \ \ \ \ \ \ \ \ \ \textbf{end for}
\STATE \ \ \ \ \ \ \ \ \ \ \ \ Compute $\textit{\textbf{u}}_i$ and $\alpha_i$ using Eq. (\ref{policy})
\STATE \ \ \ \ \ \ \ \textbf{end for}
\STATE \textbf{end for}
\end{algorithmic}
\end{algorithm}

\subsection{Reward design}
To balance the control performance and communication savings, we designed the reward of agent $i$, given as follows:
\begin{eqnarray}
r_i&=&-\| \textit{\textbf{e}}\|_2-\lambda_1 \sum_{j\in \mathcal{N}_i}\gamma_{ij}-\lambda_2 p_i \label{reward} \\
p_i&=&
\begin{cases}
1, \ \rm{if \ } \| \textit{\textbf{x}}\it{}_i-\textit{\textbf{x}}\|_{\rm{2}}>\delta \\
0, \ \rm{otherwise}\\
\end{cases}
\label{penalty}
\end{eqnarray}
where $\delta$ and $\lambda_i$ ($i=1, 2$) are the positive constant and hyperparameters, respectively.
The second term in Eq. (\ref{reward}) minimizes the communication with neighboring agents at every control step.
Moreover, we added the third term in Eq. (\ref{reward}) to improve the learning efficiency by making agents move within a certain distance $\delta$ from the payload's position.

\subsection{Policy optimization}
The weight parameters in the policy networks in the learning process are optimized to maximize the expected reward.
In this study, we optimized the multi-agent policies using the multi-agent deep deterministic policy gradient (MADDPG) \cite{Lowe2017}, which is a multi-agent variant of the deep actor-critic algorithms.

One of the primary problems in MARL is that the variance of the policy gradient can be large when the number of agents increases in partially
observable environments. To address this issue, the MADDPG algorithm adopts a learning framework called "centralized training and
decentralized execution." The critic networks approximate the optimal Q-value functions using observations and actions of all agents.
In contrast, the policy networks are optimized using a policy gradient method, wherein each actor network can access its own observations and
actions. Once the policies are trained, each policy network can compute the action under local observations.
Details of the policy optimization steps can be found in \cite{Lowe2017}.

\section{Simulation}
We demonstrate the effectiveness of the proposed algorithm through cooperative transport and cooperative rotation tasks in a simulation.
We confirmed the versality of the framework through a cooperative transport task using two agents for randomly arranged initial
and the desired positions of the payload.
Moreover, we confirmed the scalability of our framework through a cooperative rotation task using varying number of agents.

\subsection{Cooperative transport task}
\subsubsection{Setup}
\begin{figure}
\begin{center}
\includegraphics[width=4.cm]{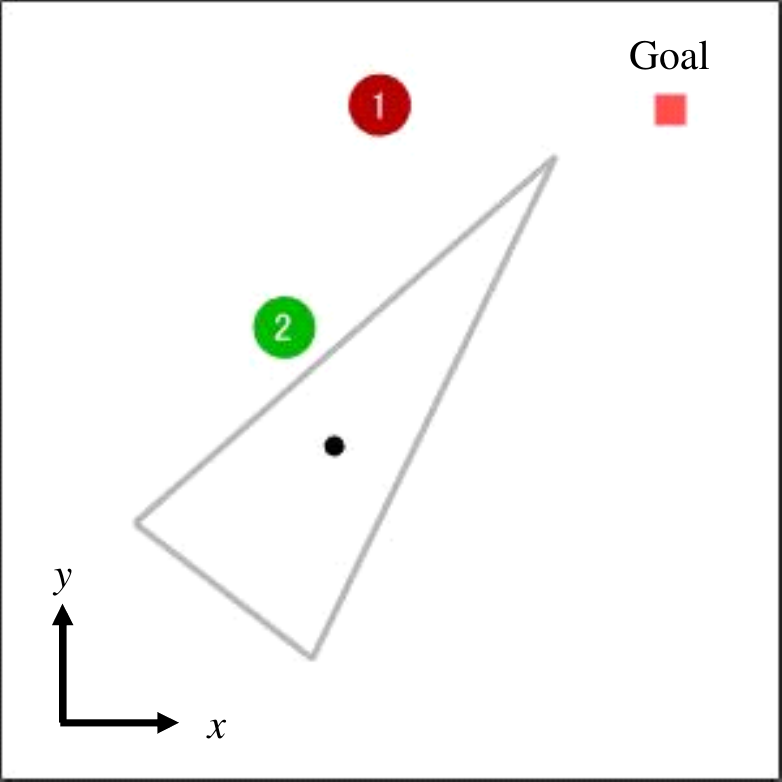}
\caption{Cooperative transport task. The colored circles, gray triangle, and black dot represent the agents, payload, and the position of the payload, respectively.}
\label{sim1env}
\end{center}
\end{figure}

We began with a 2D cooperative transport task to confirm that the proposed algorithm could balance the control performance and communication saving for randomly generated initial positions and the desired positions of the payload, as shown in Fig. \ref{sim1env}.
We used a triangular object with side lengths of 1.0, 1.0, and 0.4 m.
The mass and moment of inertia were set to 1.0$\times10^{1}$ kg and $6.0\times10^{-2}$ kg$\cdot \rm{}m^2$, respectively.
The shape of the agent was circular. The radius, mass, and moment of inertia were set to 0.10 m, 1.1 kg and 5.3 kg$\cdot \rm{}m^2$, respectively.
The control input of agent $i$ was $\textit{\textbf{u}}_i=[u^v_i, u^{\omega}_i]^{\top}$, where $u^v_i\in \mathbb{R}$ and $u^{\omega}_i\in \mathbb{R}$ are the linear and angular velocity inputs, respectively.
We also set $\mid u^v_i\mid \le 0.4$ and $\mid u^{\omega}_i\mid \le 2.0$.

We set the number of agents and neighborhood agents to $N=2$ and $K=1$, respectively.
The initial positions of the payload and agents were randomly generated within region $Q:=\{(x,y)\mid2.0\le x\le 3.0, 2.0\le y\le 3.0\}$, whereas the desired position of the payload was randomly generated within region $Q$.

Further, numerical simulations were carried out using the code in \cite{MADDPGcode} and the dynamics presented in \cite{WangDARS2016}.
Table \ref{sim1condition} lists the simulation conditions.
The parameters used in the MARL method were set by trial and error.

To confirm the effectiveness of the proposed framework, we compared our algorithm with two communication topologies, as follows:
\begin{itemize}
\item \textbf{Full}: each agent receives the resultant force and torque at every control step.
\item \textbf{Nocom}: each agent never receives force and torque from other agents.
\end{itemize}

We carried out three trainings for each communication topology under the same conditions.

Moreover, to evaluate each communication topology quantitatively, we defined the control error and communication cost as follows:
\begin{eqnarray}
E&=&\frac{1}{M}\sum^M_{m=1}\| \textit{\textbf{e}}_m(T)\|_2 \label{control} \\
C&=&\frac{1}{M}\sum^M_{m=1}\sum^N_{i=1}\sum^N_{j=1}\sum^T_{k=1}\gamma^m_{ij}(k)\frac{\Delta T}{\Delta t} \label{com}
\end{eqnarray}
where $M$ is the number of trials. We set $M=1.0\times 10^3$ in the evaluations.

\begin{table}
\caption{Simulation conditions of the cooperative transport task}
\label{sim_condition}
\centering
\renewcommand{\arraystretch}{1.1}
\begin{tabular}{cc}
\hline
Variable & Value \\
\hline
Control period [s] & 0.1 \\
Consensus period [s] & 0.05 \\
$k$ in Eq. (\ref{consensus}) & 0.5 \\
Number of steps per episode & 1.5$\times10^{2}$ \\
Number of episode & 8.0$\times10^{5}$ \\
Number of hidden layers (critic) & 4 \\
Number of hidden layers (actor) & 4 \\
Number of units per layer & 64 \\
Activation function of hidden layers & ReLU \\
Activation function of output layers (critic) & linear \\
Activation function of output layers (actor) & tanh \\
Discount factor & 0.99 \\
Batch size & 4096 \\
Replay buffer & 1.0$\times10^6$ \\
\hline
\end{tabular}
\label{sim1condition}
\end{table}

\subsubsection{Training performance}
\begin{figure}
\begin{center}
\includegraphics[width=7cm]{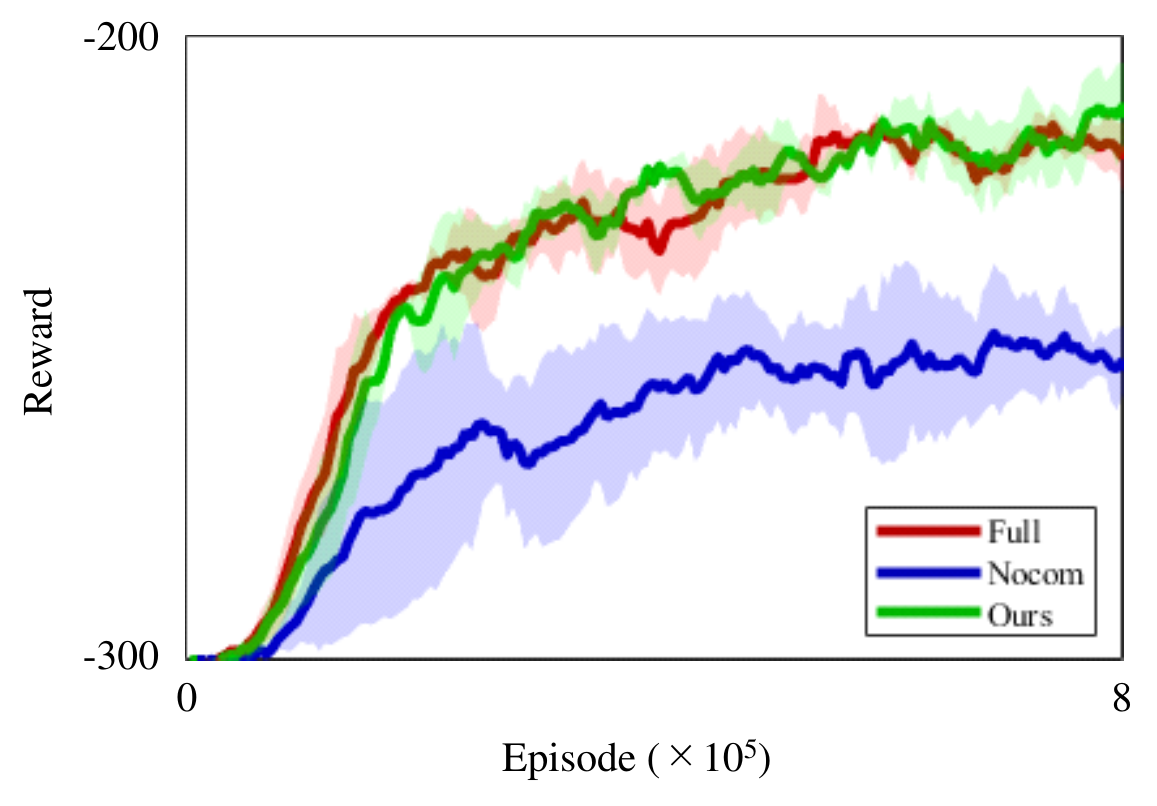}
\caption{Comparison of the cumulative reward of the first term in Eq. (\ref{reward}) when applying each method to the cooperative transport task.}
\label{reward1sim1}
\end{center}
\end{figure}

Figure \ref{reward1sim1} shows the comparison of the cumulative reward of the first term in (\ref{reward}) at each episode when applying each method to the cooperative transport task.
The results showed that the proposed method achieved a cumulative reward as high as that of the \textbf{Full} method.
Moreover, it was greater than that of the \textbf{Nocom} method.
These results indicate that the estimates of the resultant force and torque could improve the cumulative reward.

\subsubsection{Transport performance and communication saving}
\begin{figure*}
\centering
\subfigure[\textbf{Full}]{
\includegraphics[width=3.5cm]{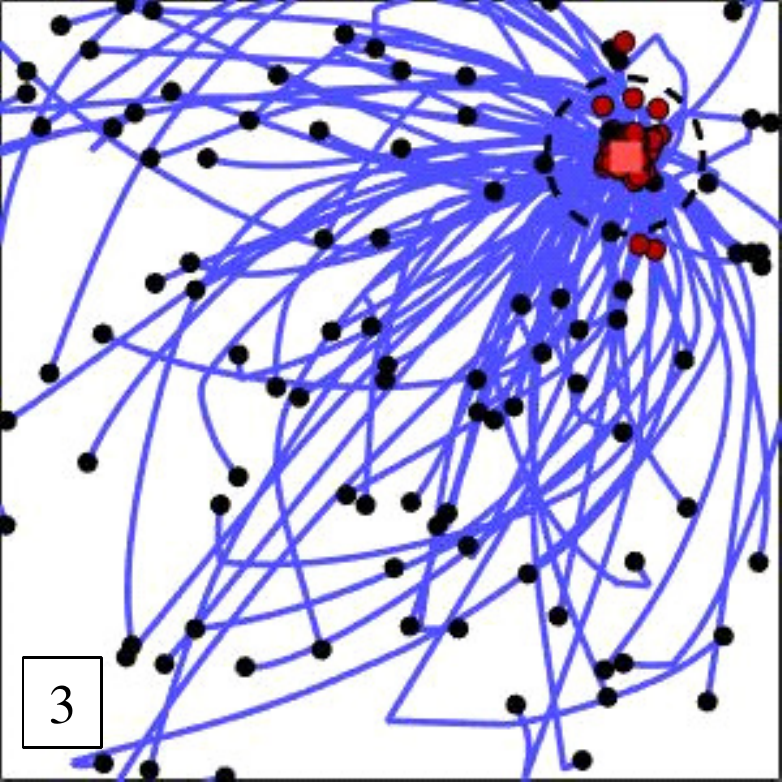}
\label{trajectory-sim1}}
\hspace{10mm}
\subfigure[\textbf{Nocom}]{
\includegraphics[width=3.5cm]{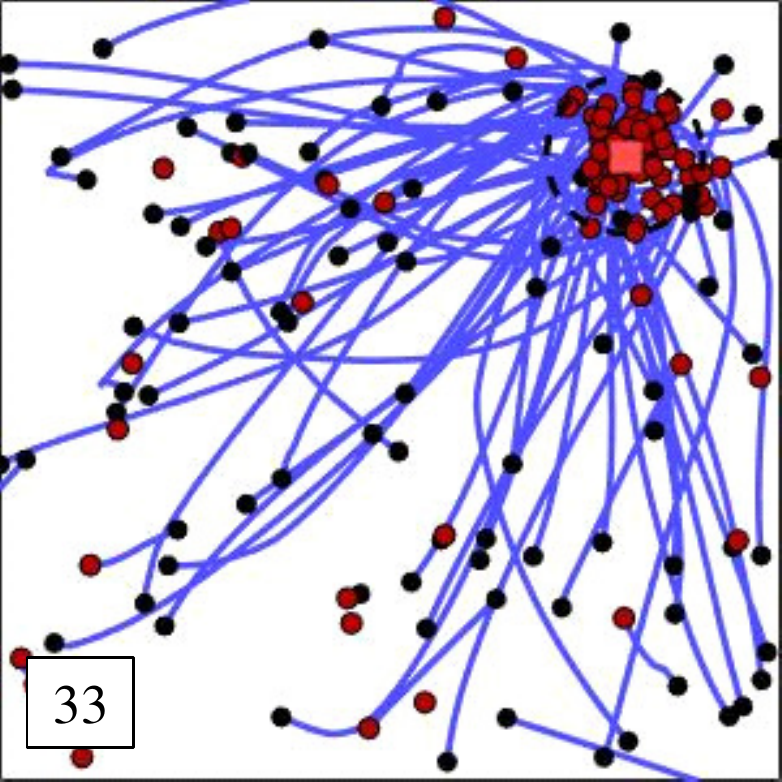}
\label{agent1}}
\hspace{10mm}
\subfigure[\textbf{Ours}]{
\includegraphics[width=3.5cm]{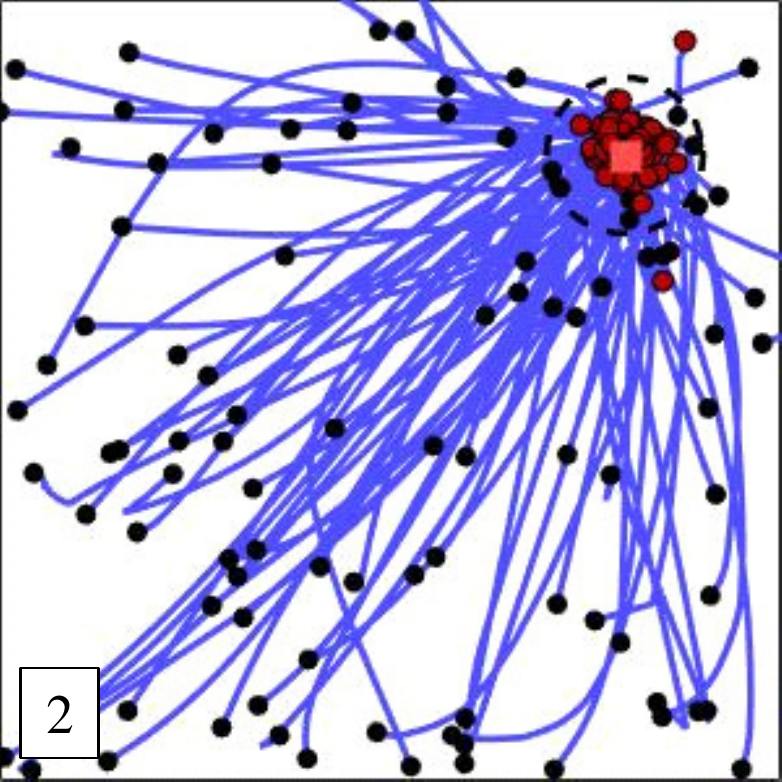}
\label{agent2}}
\caption{Results of position control for randomly arranged initial positions of the payload when applying each method to the cooperative
transport task. The blue lines, black dots, red dots, and dashed circles denote the trajectories of the payload, positions of the payload at the
initial and last steps, and the areas within 0.1 m from the desired position, respectively. The number denotes the total number of trials where the payload fails to be controlled within 0.1 m from the desired position.}
\label{versality}
\end{figure*}

Figure \ref{versality} shows the results of cooperative transport for randomly arranged initial positions of the payload when each method is
applied. We performed 100 trials for each method.
The results showed that the \textbf{Full} and \textbf{Ours} successfully controlled the position of the payload within 0.1 m from the desired
position for most trials, whereas the \textbf{Nocom} method failed for 33 trials.

Table \ref{sim1result} shows the comparisons of the mean absolute error and communication cost for each communication topology.
The results showed that our method achieved errors as small as that of the \textbf{Full} method.
Moreover, compared to the \textbf{Full} method, our method saved communication costs by $63\%$.

Overall, our method achieved transport performance as good as that of full communication topology while reducing communication costs for randomly generated initial positions and desired positions of the payload.

\subsubsection{Communication and estimation}
Herein, we verify that our method can determine communication timing while estimating the resultant force and torque.
Figure \ref{trajectory-sim1} shows the trajectories and communication occurrence at each time.
The results showed that communication occurred while agents determined the edge that they should push.
During this period, agent 1 changed the pushing position on the same edge as agent 2 to move the payload.
Afterwards, the agents kept pushing the payload without communication until the payload reached the desired position.

Figures \ref{agent1} and \ref{agent2} show the communication occurrences and the estimates of the resultant force and torque by two agents.
The results showed that the communication rate was high a few seconds after the start of the control, whereas it was low at the end of the control.
Moreover, our method estimated the resultant force and torque closer to the true values, compared to the method without consensus when communication occurred in the red-colored regions.

Overall, our method could determine the timing of communication of neighborhood agents while estimating the resultant force and torque using communication.

\begin{table}
\caption{Comparisons of the control error and communication cost when applying each method to the cooperative transport task.}
\centering
\renewcommand{\arraystretch}{1.1}
\begin{tabular}{cccc}
\hline
& \textbf{Full} & \textbf{Nocom} & \textbf{Ours} \\
\hline
$E$ & 0.05 m & 0.22 m & \textbf{0.05} m\\
$C$ & 4.0$\times10^2$ & 0.0 & \textbf{1.5$\times10^2$} \\
\hline    
\end{tabular}
\label{sim1result}
\end{table}

\begin{figure*}
\centering
\subfigure[Trajectories and communication occurence]{
\includegraphics[width=15cm]{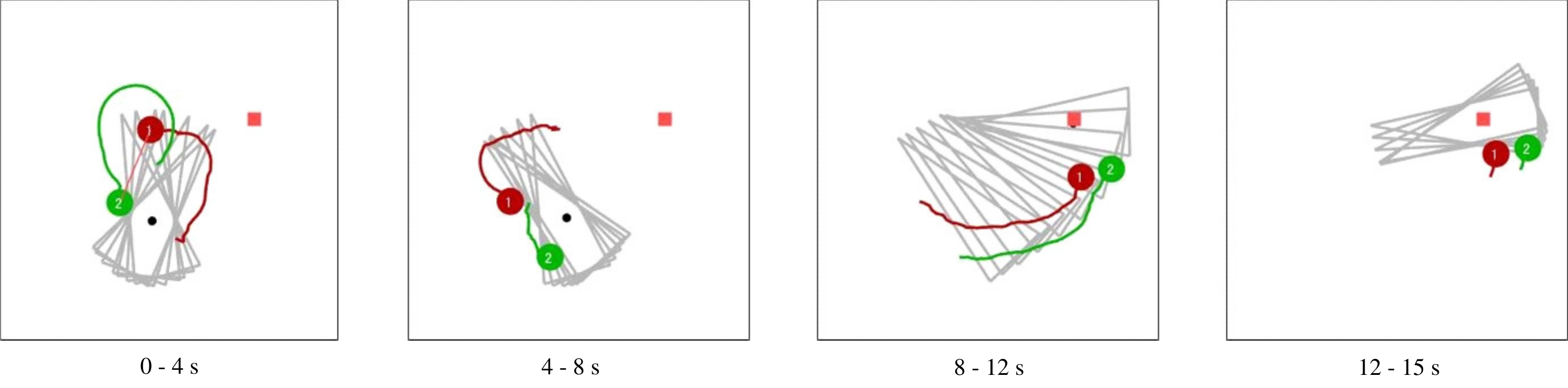}
\label{trajectory-sim1}}
\subfigure[Agent 1]{
\includegraphics[width=16cm]{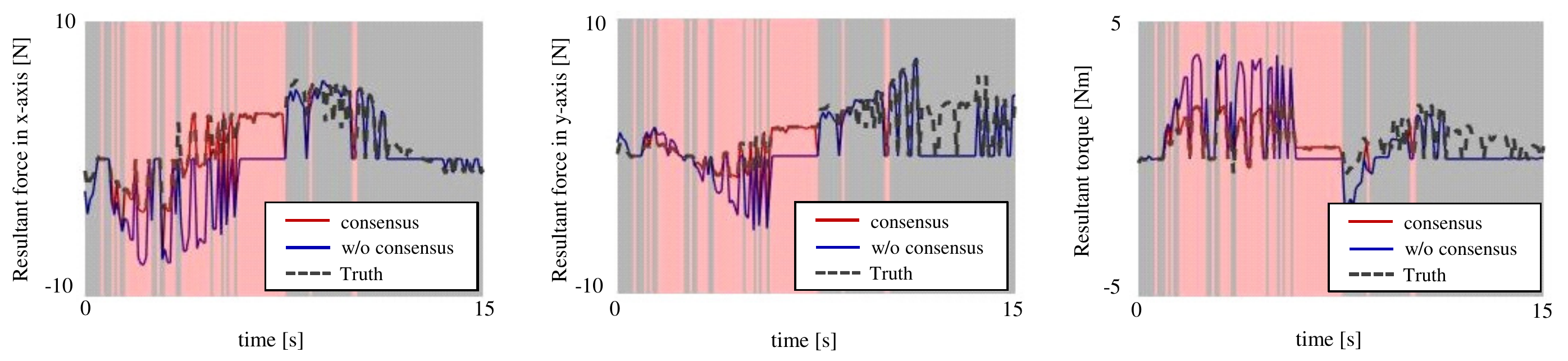}
\label{agent1}}
\subfigure[Agent 2]{
\includegraphics[width=16cm]{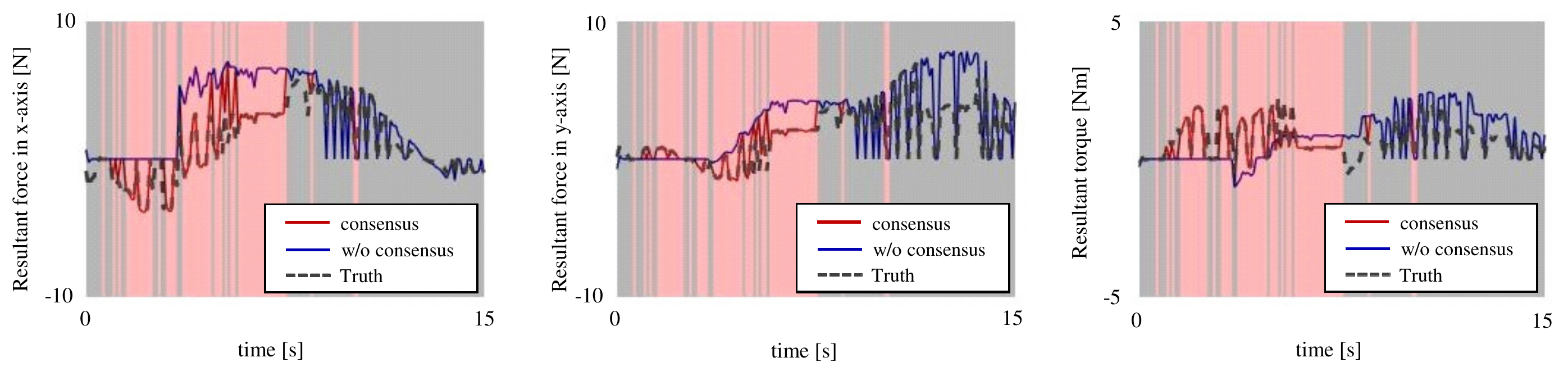}
\label{agent2}}
\caption{Trajectories, communication occurrence, and the estimation results when applying the proposed method to the cooperative transport task. The red lines in Fig. \ref{trajectory-sim1} indicate that mutual communications of the estimated resultant force and torque occur between agents.
In Figs. \ref{agent1} and \ref{agent2}, each agent receives the resultant force and torque estimates from other agents in red-colored regions. In contrast, the agents do not receive them in gray-colored regions.}
\label{sim1}
\end{figure*}

\subsection{Cooperative rotation task}
\subsubsection{Setup}
\begin{figure}
\begin{center}
\includegraphics[width=4.5cm]{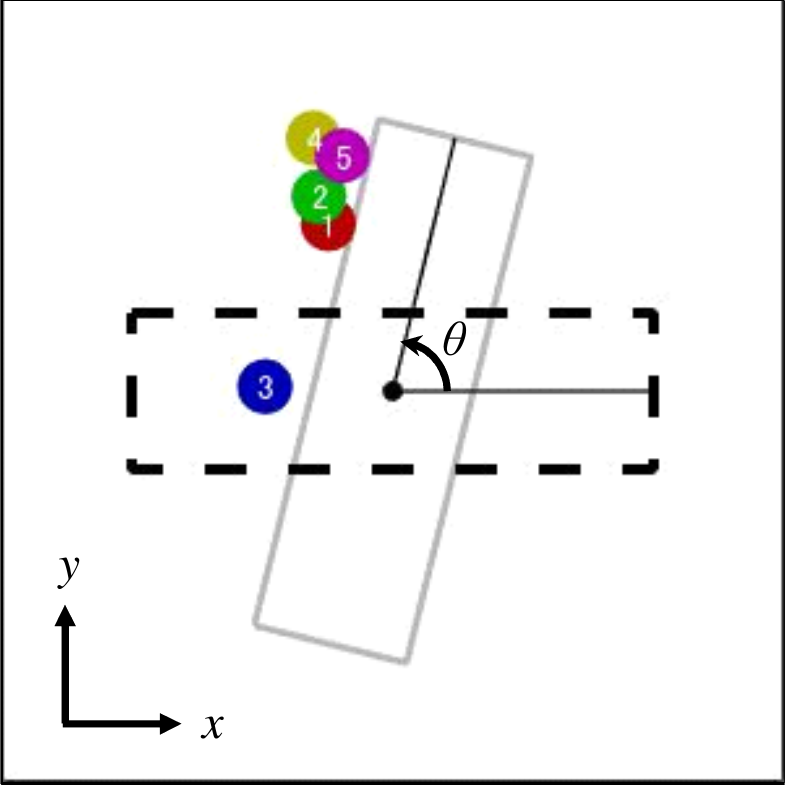}
\caption{Cooperative rotation task. The colored circles, gray rectangular, and dashed rectangular denote the agents, payload, and the desired state of the payload, respectively.}
\label{sim2env}
\end{center}
\end{figure}

The second simulation is a cooperative rotation task to confirm that our algorithm can balance control performance and communication savings for varying number of agents, as shown in Fig. \ref{sim2env}.

The mass and moment of inertia were set to 2.0$\times 10^1$ kg and 7.3 kg$\cdot \rm{}m^2$, respectively.
The shape, radius, mass, and moment of inertia were set to the same value as in the previous simulation.
Furthermore, we set $\mid u^v_i\mid \le 0.2$ and $\mid u^{\omega}_i\mid \le 2.0$.

The yaw angle in the world coordinate is defined as shown in Fig. \ref{sim2env}.
The center position of the payload is fixed to [2.0 m, 2.0 m]$^{\top}$.
The initial yaw angle of the payload was randomly generated within $[0.4\pi, 0.6\pi]^{\top}$, whereas the desired yaw angle was set to 0 or $\pi$.
The initial positions of the agents were randomly set to [1.35 m, 1.0 m]$^{\top}$ or [1.35 m, 3.0 m]$^{\top}$ with a 50$\%$ probability.

The simulation conditions are listed in Table. \ref{sim2condition}.
The remaining parameters were set to similar values as in the previous simulation.
We set the number of agents and neighborhood agents to $N=5$ and $K=2$, respectively, in the training phase.
To confirm the scalability of our method, we executed the trained policies for $N\in\{3, 4, 5, 6, 7, 8\}$.
Moreover, we set the parameters in Eqs. (\ref{reward}) and (\ref{penalty}) to $\lambda_1=0.02$, $\lambda_2=0.1$ and $\delta=1.2$ by trial and error, respectively.

To confirm the effectiveness of our algorithm, we compared our algorithm with three communication topologies as follows:
\begin{itemize}
\item \textbf{Full}: each agent receives the resultant force and torque at every control step.
\item \textbf{ETC}: each agent determines the timing for agents to receive the force and torque using the event-triggered communication
\cite{Shibata2021}.
\item \textbf{Nocom}: no agent receives force and torque from any other agents.
\end{itemize}

We carried out three trainings for each communication topology under the same conditions.

Moreover, we evaluated the control error $E$, communication cost $C$, and transport time $P$, which were defined by the average time to control the error of the yaw angle within $\pm 10$ deg for $1.0\times 10^3$ trials.

\begin{table}
\caption{Simulation conditions of the cooperative rotation task}
\centering
\renewcommand{\arraystretch}{1.1}
\begin{tabular}{cc}
\hline
Variable & Value \\
\hline
Control period [s] & 0.1 \\
Consensus period [s] & 0.02 \\
$k$ used in the consensus & 0.2 \\
Number of steps per episode & 2.0$\times10^{2}$ \\
Number of episode & 6.0$\times10^{5}$ \\
\hline
\end{tabular}
\label{sim2condition}
\end{table}

\subsubsection{Training performance}
\begin{figure}
\begin{center}
\includegraphics[width=7cm]{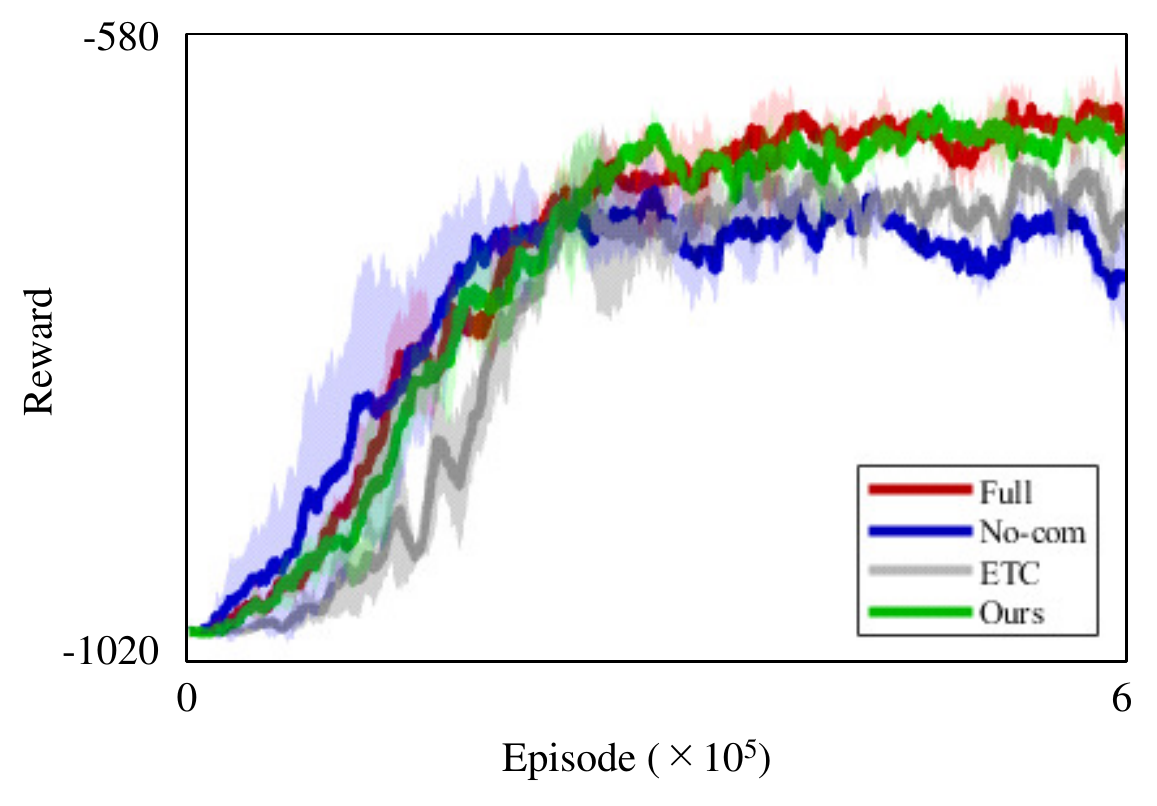}
\caption{Comparisons of the cumulative reward of the first term in Eq. (\ref{reward}) when applying each method to the cooperative rotation task.}
\label{reward1sim2}
\end{center}
\end{figure}

Figure \ref{reward1sim2} compares the cumulative reward of the first term in Eq. (\ref{reward}) when applying each method to the cooperative
rotation task.
The results showed that our method achieved cumulative rewards as high as that of the \textbf{Full} method.
Moreover, the rewards were greater than those of the \textbf{ETC} and \textbf{Nocom} methods.
These results show that the estimates of the resultant force and torque improved the cumulative reward.

\subsubsection{Scalability analysis}
\begin{figure}
\centering
\subfigure[Control error]{
\includegraphics[width=7.5cm]{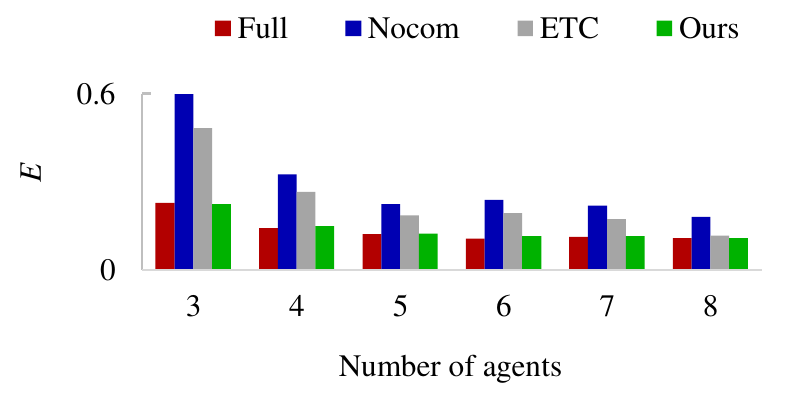}
\label{errorsim2}}
\subfigure[Transport time]{
\includegraphics[width=7.5cm]{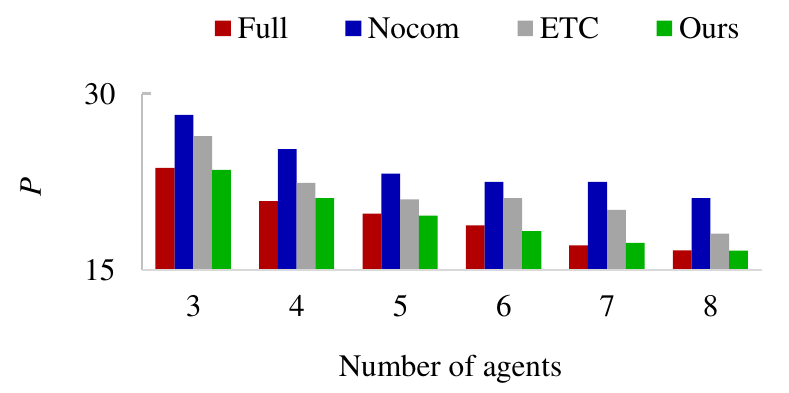}
\label{timesim2}}
\subfigure[Communication cost]{
\includegraphics[width=7.5cm]{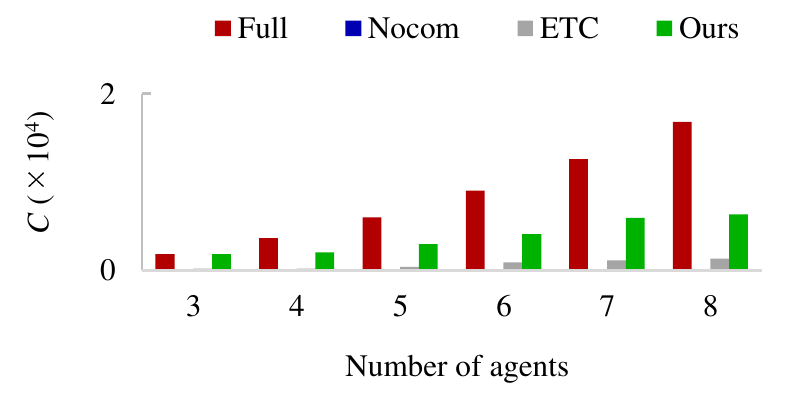}
\label{comsim2}}
\caption{Comparison of the control error, transport time, and communication cost when applying each method to the cooperative rotation using various number of agents.}
\label{scalability}
\end{figure}

Figure \ref{errorsim2} compares the control errors when applying each method for varying number of agents.
The results showed that the control errors of the \textbf{Nocom} and \textbf{ETC} methods became considerably large when the number of agents was small.
This can be attributed to the fact that the yaw angle could not reach the desired value in the given control steps, considering the resultant torque decreased as the number of agents decreased.
In contrast, the \textbf{Full} and our methods reduced these errors compared to the \textbf{Nocom} and \textbf{ETC} methods.
As the number of agents increased, the differences among the \textbf{Full}, \textbf{ETC}, and proposed methods decreased.
Meanwhile, the \textbf{Full} and proposed methods achieved shorter transport time compared to the \textbf{ETC} method, as shown in Fig. \ref{timesim2}.

Figure \ref{comsim2} compares the communication costs.
The results showed that the communication cost of the proposed method was higher than that of the \textbf{ETC} method, considering that our method required high-rate communication to estimate the resultant force and torque.
Meanwhile, our method decreased the communication cost significantly compared to the \textbf{Full} communication, whose control performance was almost as good as that of our method.

Overall, our method achieved transport performance as good as that of the \textbf{Full} communication topology while saving the communication costs for varying number of agents.

\subsubsection{Communication and estimation}
\begin{figure*}
\centering
\subfigure[Trajectories and communication occurence]{
\includegraphics[width=14cm]{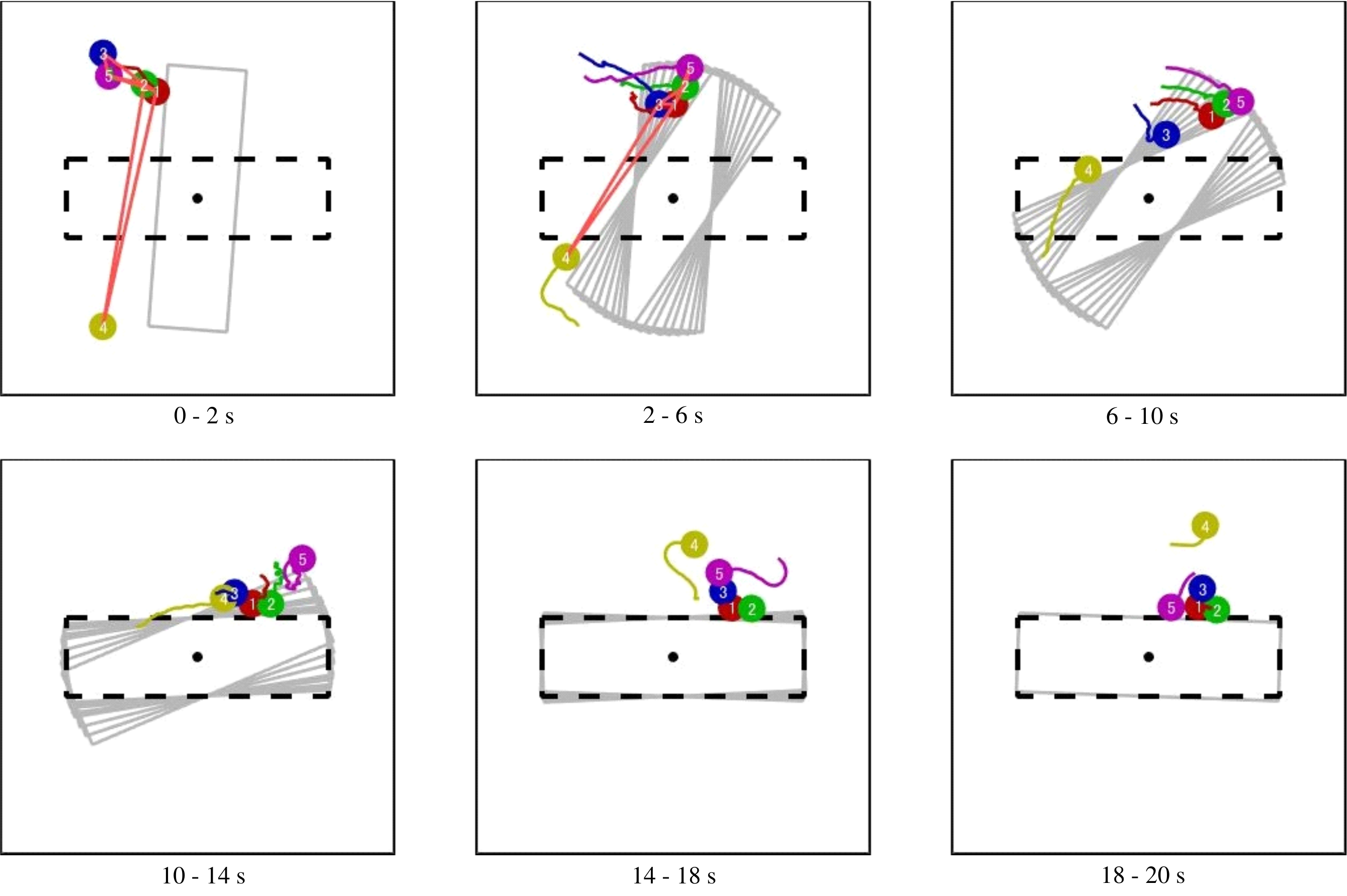}
\label{trajectory}}
\vspace{2mm}
\subfigure[Resultant torque estimated by each agent]{
\includegraphics[width=17cm]{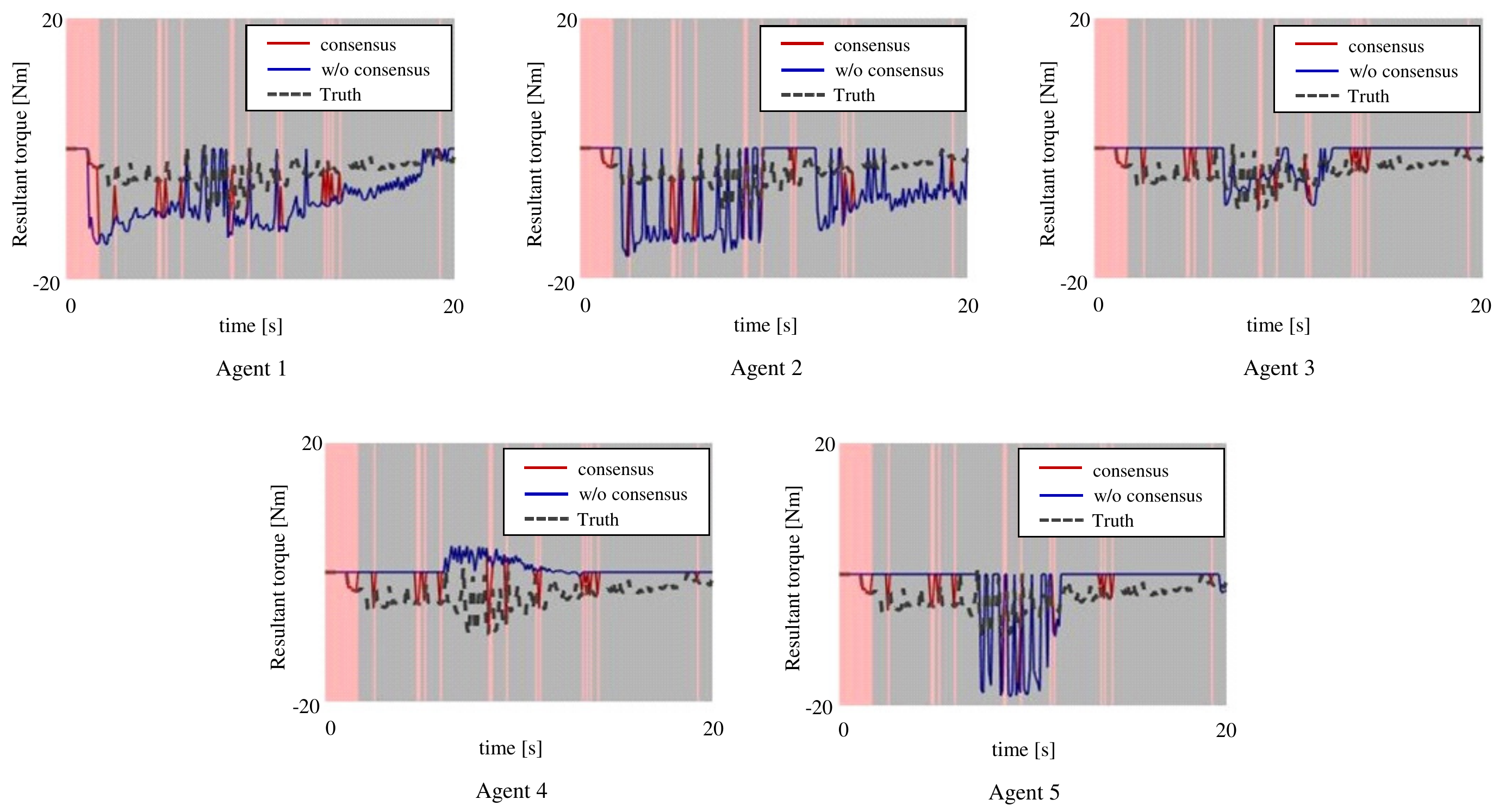}
\label{estimation}}
\caption{Trajectories, communication occurrence, and the estimation results when the proposed method is applied to the cooperative rotation task.
The red lines in Fig. \ref{trajectory} show that mutual communications of the estimated resultant force and torque occur between agents. In Fig. \ref{estimation}, each agent receives the resultant force and torque estimates from other agents in red-colored regions. In contrast, the agents do not receive them in gray-colored regions.}
\label{sim2}
\end{figure*}

In this subsection, we verify that our method can determine the communication timing while estimating the resultant torque using local observations.
Figure \ref{trajectory} shows the trajectories and communication occurrence at each time.
The results showed that communication occurred while agents determine the direction in which they rotate the payload.
During this period, agent 4 changed the pushing position to rotate the payload in the same direction as the other agents.
Following this, agents kept pushing the payload until the yaw angle of the payload reached the desired value.

Figure \ref{estimation} depicts the communication occurrence and the estimates of the resultant torque by all agents.
The agents adopted high-rate communication to determine the direction in which they rotate the payload a few seconds after the start of the
control.
Once the payload rotated clockwise, they adopted low-rate communication, considering that they pushed the payload until the yaw angle of the payload
reached the desired value.
Moreover, our method estimated the resultant force and torque closer to the true values compared to those of the method without consensus when communication occurred in red-colored regions.

Overall, the proposed method could determine communication timing with the neighboring agents while estimating the resultant torque using local observations.

\subsubsection{Connectivity of communication topology}
Next, we verify the manner in which a communication topology using our method may satisfy the first condition of the average consensus by introducing a
connectivity metric $R_c$, as described in Appendix.

Table \ref{connectivity} shows a comparison of $R_c$ for each method. We compared our method with a \textbf{Random} method, where we randomly skipped communication among agents with a probability of 50$\%$.
The proposed method improves the estimation accuracy if the communication topology satisfies the connectivity condition.
Moreover, the estimation accuracy improves the transport performance.
Consequently, our method achieved a value of $R_c=0.7$ although we did not promote the connectivity in the reward.
In contrast, the $\textbf{Nocom}$, $\textbf{ETC}$, and $\textbf{Random}$ methods resulted in $R_c$ values that were lower than the value of our method as they did not require the connectivity condition.
These results show that our method could obtain the necessary condition to estimate the resultant force and torque using the estimates of the resultant force and the torque of the neighborhood agents.

\begin{table}
\caption{The ratio of the connectivity}
\centering
\renewcommand{\arraystretch}{1.2}
\begin{tabular}{ccccc}
\hline
& \textbf{Nocom} &\textbf{ETC} & \textbf{Random} & \textbf{Ours} \\
\hline
$R_c$ & 0.0 & 0.0 & 0.23 & \textbf{0.7}\\
\hline    
\end{tabular}
\label{connectivity}
\end{table}

\section{Real robot experiment}
Herein, we show the effectiveness of our learning method through real robot experiments using multiple ground robots to confirm that the settings in the simulation are realistic and the scalability of our method holds true in real robot environments.

\subsection{Setup}
\begin{figure}
\centering
\subfigure[Robot]{
\includegraphics[width=6cm]{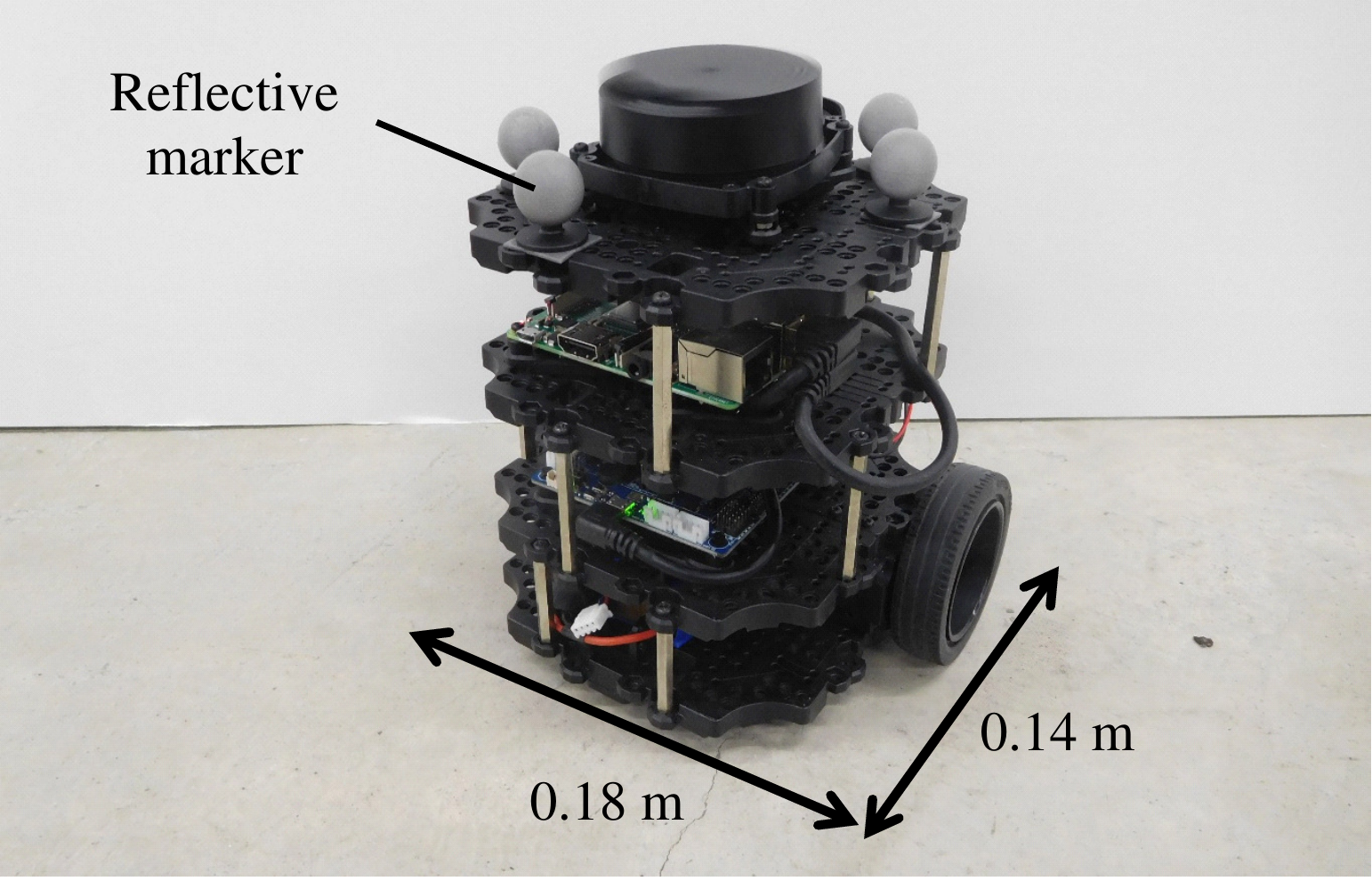}
\label{robot}}
\subfigure[Experimental configuration]{
\includegraphics[width=6cm]{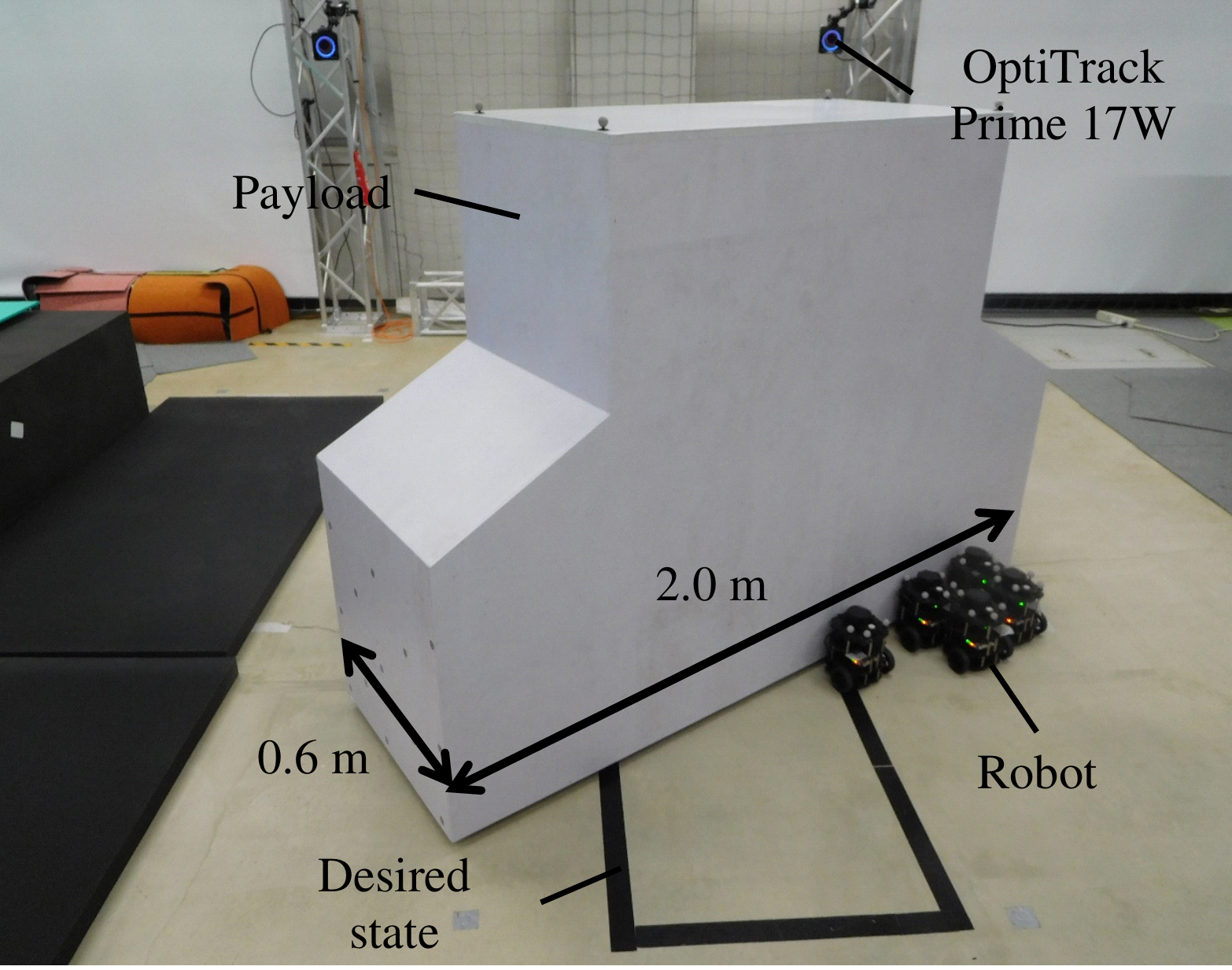}
\label{config}}
\caption{Robot and experimental configuration used in the experiment.
The positions and yaw angles of the payload and robots were observed using a motion capture system.}
\label{expsys}
\end{figure}

The real robot demonstration was performed using Turtlebot3 Burger robots, and the experimental configuration is shown in Figs. \ref{robot} and \ref{config}.

Our experimental system utilized an OptiTrack Prime 17~W motion capture system (Natural Point, Inc., Corvallis, OR) to observe the positions and yaw angles of the payload and robots at 10~Hz.
The linear and angular velocities of the payload and robots were calculated using the measured positions and yaw angles.
We trained the policies by setting the number of robots and neighborhood robots to $N=5$ and $K=2$, respectively, in the simulation and executed the trained policies for $N\in\{3,4,5,6\}$ in the real environment.
The trained policies calculated the control inputs in a PC with an 8-core Intel{\textregistered} Core{\texttrademark} i7 (2.80~GHz) with 32~GB RAM.
The control inputs were transmitted from the control PC to each robot using Wi-Fi communication at 10~Hz.

\subsection{Result}
\begin{figure}
\begin{center}
\includegraphics[width=7.5cm]{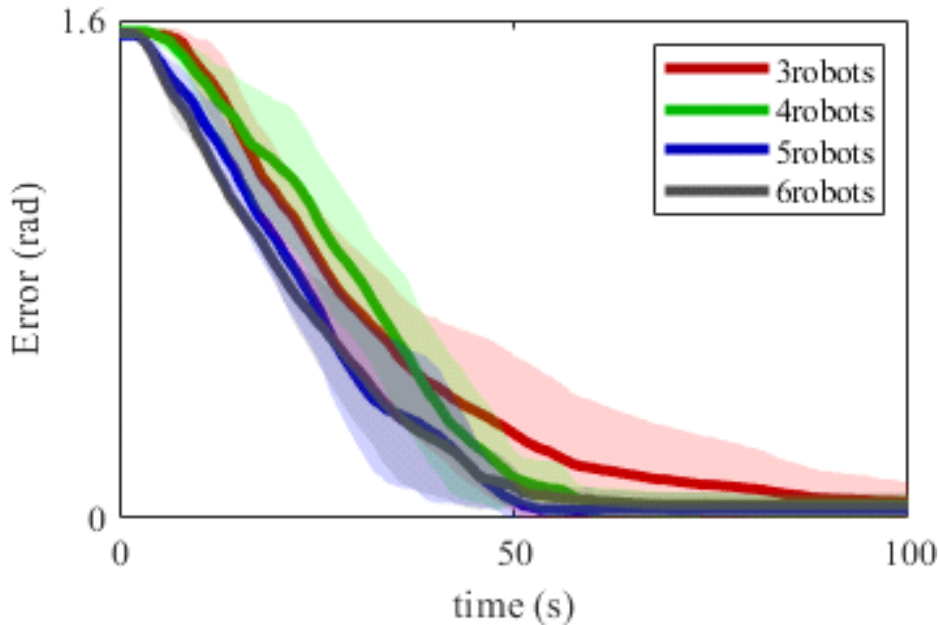}
\caption{Mean absolute error of the yaw angle when applying our method to various number of robots for 10 trials.}
\label{eval}
\end{center}
\end{figure}

Figure \ref{eval} shows the mean absolute error of the yaw angle when applying our method to various numbers of robots.
The results show that our method can control the yaw angle of the payload to the desired value for various numbers of robots.
Moreover, the time for controlling the yaw angle becomes shorter as the number of robots increases.

\begin{figure*}
\centering
\subfigure[Three robots]{
\includegraphics[width=17cm]{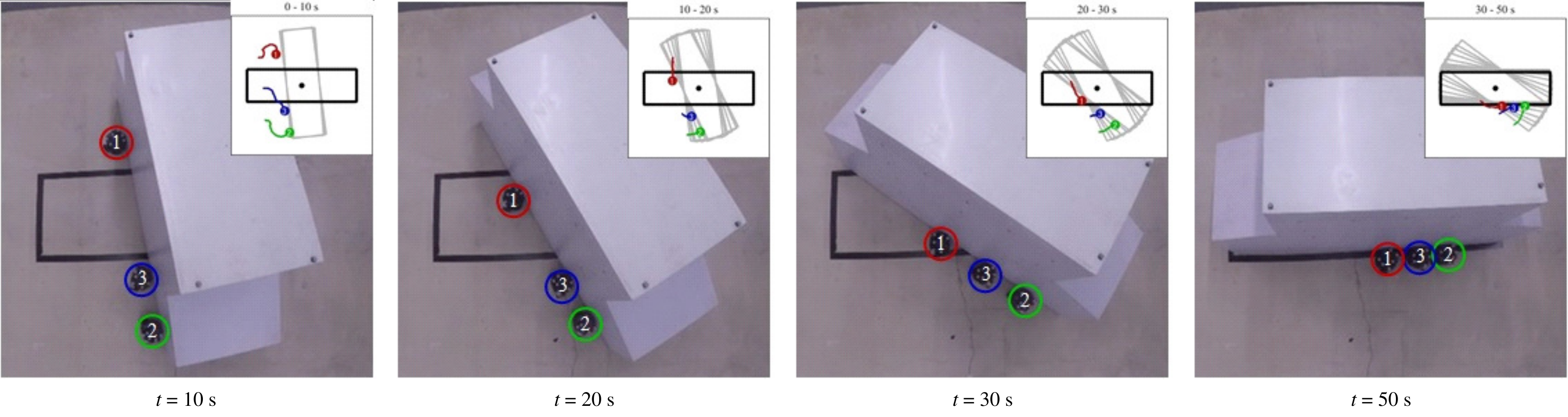}
\label{3robot}}
\subfigure[Four robots]{
\includegraphics[width=17cm]{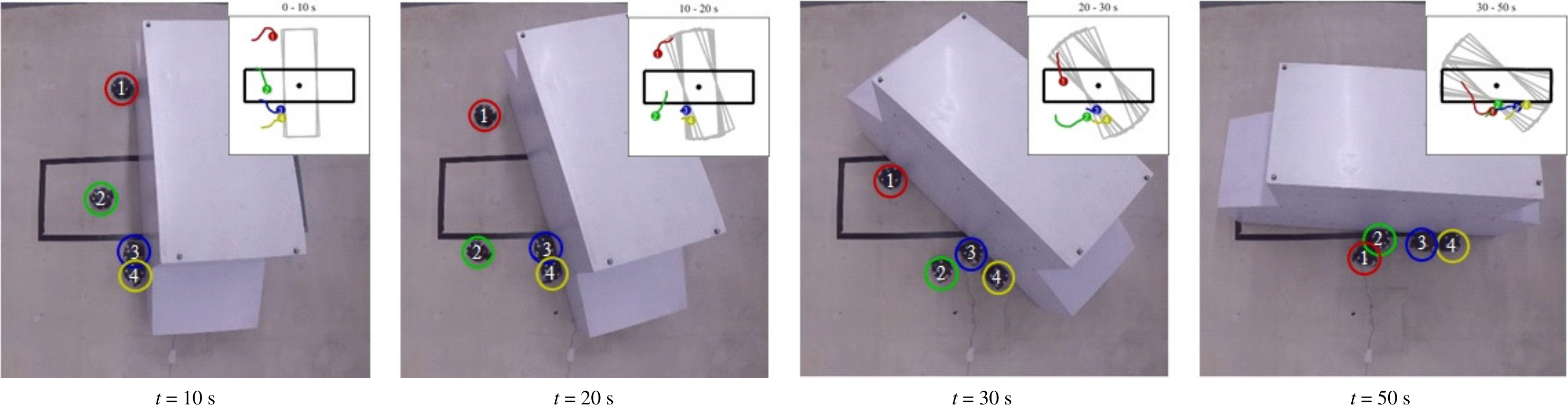}
\label{4robot}}
\subfigure[Five robots]{
\includegraphics[width=17cm]{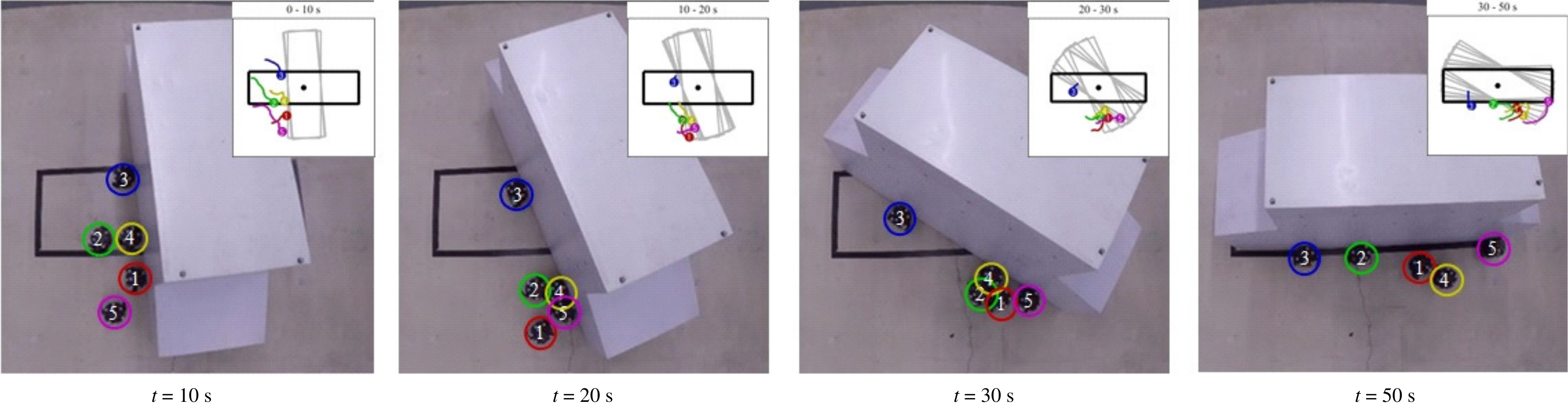}
\label{5robot}}
\subfigure[Six robots]{
\includegraphics[width=17cm]{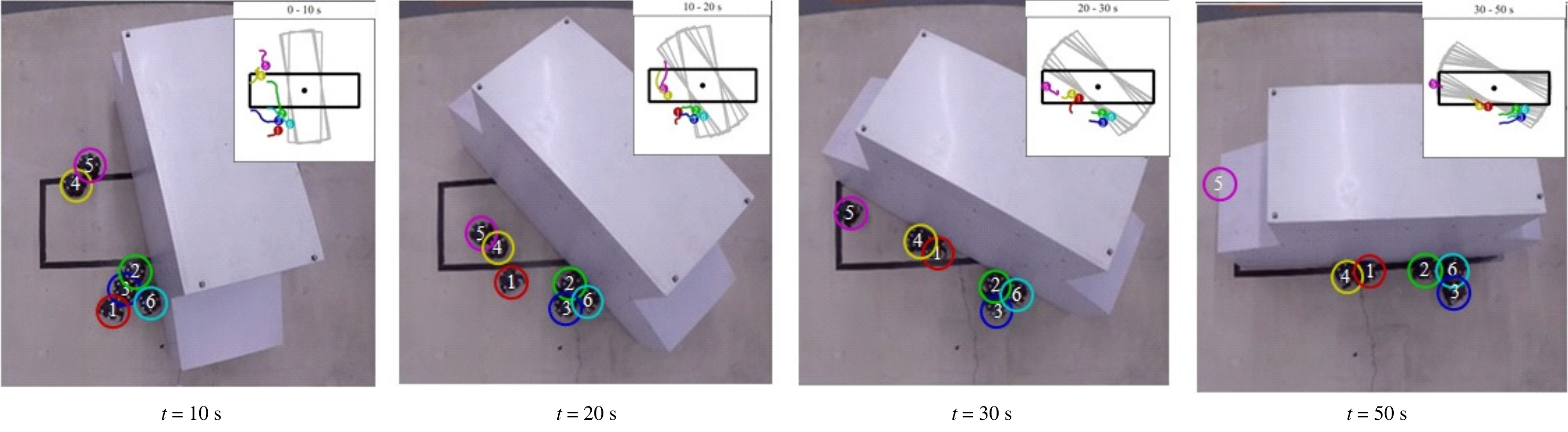}
\label{6robot}}
\caption{Trajectories of the payload and robots when applying our method to various number of robots. We also show the top view of the trajectories of the payload and robots. The black line denotes the desired state of the payload.}
\label{demo}
\end{figure*}

Figure \ref{demo} depicts the trajectories of the payload and robots for various number of robots.
After several robots came in contact with the payload, the payload began to rotate clockwise.
Once the payload rotated, several robots changed the pushing positions to rotate the payload in the same direction as the other robots.
The robots kept rotating the payload collaboratively until the yaw angle of the payload reached the desired value for various number of robots.

Overall, we verified that the settings in the simulation were realistic and the scalability of our method holds true in real robot environments.

\section{Discussion}
In what follows, we discuss the satisfaction of consensus conditions.
As presented in~\cite{Kennedy2015}, the communication topology is connected if each robot communicates with robots on both sides.
However, such an assumption is not always realistic, considering that it is difficult for robots at both ends to communicate with each other.
Although our method cannot theoretically guarantee the graph's connectivity, it achieves higher connectivity than other communication topologies.

Through simulations, it has been confirmed that the proposed method can tolerate a certain degree of fluctuations in the number of agents. However, there is room for improvement when increasing the number of agents. Because our method cannot theoretically guarantee the connectivity of communication topology, the connectivity may decrease when the number of agents increases. To address this issue, we will combine the proposed method with a learning method to maximize the connectivity achieved in a previous study~\cite{Daniel2022}. Furthermore, the convergence speed decreases as the number of agents increases~\cite{Olshevsky2009}, which can degrade control performance. Therefore, it is crucial to consider a framework that guarantees a certain level of convergence speed.

In this study, the trained multi-agent policies performed worse in real experiments as the number of robots increased, considering that our model neglected the collision avoidance between different robots.
To address this issue, we plan on combining our algorithm with decentralized
multi-robot collision avoidance presented in~\cite{Zhai}.

Because we assume mutual communication among agents in the communication laws, our method requires a few modifications to implement Eq. (\ref{update}). Let us consider a situation where agent $j$ decides to communicate with agent $i$. In this case, agent $j$ sends a signal to communicate with agent $i$. When agent $i$ observes this signal, agent $i$ decides to communicate with agent $j$ in the next consensus step using Eq. (\ref{update}). Thereafter, agents $i$ and $j$ can communicate with each other.

Reducing the sample complexity in the current system would be an important line of research.
While we could optimize the multi-agent policies for a small number of robots, the computational costs could significantly increase as the number of robots increases.
To address this issue, we will combine our algorithm with a more sample efficient algorithm such as a multi-agent model-based RL algorithm \cite{Willemsen}.

It would be interesting to apply the proposed method to a three-dimensional cooperative transport task. Our algorithm can estimate the resultant force and torque in three dimensions using Eq. (\ref{consensus}). However, to calculate the three-dimensional resultant torque, it is necessary to know the object's moment of inertia. To address this problem, we will combine the proposed method with the moment of inertia estimation technique proposed in~\cite{Franchi2015}.

\section{Concluson}
In this paper, we proposed a learning framework of ETC and consensus-based control for distributed cooperative
transport.
The proposed method achieved transport performance as good as that of full communication while saving the communication costs through cooperative transport tasks using two agents for randomly arranged initial and desired positions of the payload.
Moreover, our method achieved transport performance as good as that of full communication while saving the communication costs through
cooperative rotation tasks in scenarios wherein the number of agents differed from that in the training environment.
In future studies, we plan to better adapt multi-agent policies for real environments and apply our algorithm to cooperative manipulation in 3D environments.












\appendix

\section{Connectivity metrics}
This appendix introduces the connectivity metrics of communication topology.

At every control step $k$, we check the satisfaction of the connectivity given by
\begin{eqnarray}
\beta(k)=
\begin{cases}
1,\ {\rm if}\ {\rm Eq.}\ (\ref{rank})\ {\rm is}\ {\rm satisfied} \\
0,\ {\rm otherwise}
\end{cases}.
\nonumber
\end{eqnarray}

Moreover, we evaluate the ratio of the connectivity given by
\begin{eqnarray}
R_c=\frac{1}{N}\sum^N_{i=1}\frac{\sum^T_{k=1}\beta(k)}{\sum^T_{k=1}b_i(k)},
\nonumber
\end{eqnarray}
where $b_i(k)=1$ if agent $i$ communicates with other agents at control step $k$; otherwise $b_i(k)=0$.

\end{document}